%% file: main.tex
\documentclass[10pt,twocolumn,letterpaper]{article}

\usepackage{wacv}              


\definecolor{wacvblue}{rgb}{0.21,0.49,0.74}
\usepackage[pagebackref,breaklinks,colorlinks,allcolors=wacvblue]{hyperref}

\usepackage{amsmath, amssymb, amsthm, mathtools}
\usepackage{bm}

\usepackage{graphicx}
\usepackage{booktabs}
\usepackage{subcaption}

\usepackage[accsupp]{axessibility}

\usepackage{tikz}
\usetikzlibrary{positioning,fit,arrows.meta,calc,decorations.pathreplacing}

\usepackage{amsfonts}
\usepackage{mathtools}
\usepackage{amsthm}

\newtheorem{example}{Example}
\newtheorem{proposition}{Proposition}

\usepackage[inline]{enumitem}

\usepackage[utf8]{inputenc}
\usepackage{nicefrac}
\usepackage{microtype}

\usepackage{float}

\input{macros}

\def\wacvPaperID{752} 
\def\confName{WACV}
\def\confYear{2027}

\title{Weakly Supervised Segmentation as Semantic-Based Regularization} 

\author{Stefano Colamonaco\\
Ku Leuven\\
Leuven, Belgium\\
{\tt\small stefano.colamonaco@kuleuven.be}
\and
Andrei-Bogdan Florea\\
Apixa\\
Leuven, Belgium\\
{\tt\small andreiflorea02@gmail.com}
\and
Jaron Maene\\
KU Leuven\\
Leuven, Belgium\\
{\tt\small jaron.maene@kuleuven.be}
}

\begin{document}
\maketitle
\input{sec/0_abstract}

\input{sec/1_intro}
\input{sec/2_related}
\input{sec/3_preliminaries}
\input{sec/4_constraints}
\input{sec/5_logic-guided}

\input{sec/6_experiments}
\input{sec/7_results}
\input{sec/8_conclusions}

{
    \small
    \bibliographystyle{ieeenat_fullname}
    \bibliography{bibliography}
}

\newpage
\appendix
\onecolumn
\input{sec/appendix}

\end{document}

%% file: macros.tex
\newcommand{\Xvars}{\ensuremath{\mathbf{X}}}
\newcommand{\xvars}{\ensuremath{\mathbf{x}}}
\newcommand{\Xvar}{\ensuremath{X}}
\newcommand{\xvar}{\ensuremath{x}}

\newcommand{\Yvars}{\ensuremath{\mathbf{Y}}}
\newcommand{\yvars}{\ensuremath{\mathbf{y}}}
\newcommand{\Yvar}{\ensuremath{Y}}
\newcommand{\yvar}{\ensuremath{y}}

\newcommand{\dom}[1]{\Omega(#1)}

%% file: sec/0_abstract.tex
\begin{abstract}
Weakly supervised semantic segmentation (WSSS) trains segmentation models from partial or coarse annotations such as bounding boxes, scribbles, or image-level tags. While recent works generate pseudo-labels with foundation models such as the Segment Anything Model (SAM), these approaches typically depend on heuristic prompt choices and offer limited ways to incorporate prior knowledge or heterogeneous labels. We address this gap by taking a neurosymbolic perspective: implementing WSSS as differentiable fuzzy logic. We unify weak annotations and domain-specific priors as continuous logical constraints which are used to fine-tune SAM. The refined foundation model then produces improved pseudo-labels, from which we train a second-stage prompt-free segmentation model. Experiments on Pascal VOC 2012 and the REFUGE2 segmentation dataset show that our logic-guided fine-tuning yields higher-quality pseudo-labels, leading to state-of-the-art WSSS accuracy that can exceed fully supervised baselines.
\end{abstract}

%% file: sec/1_intro.tex
\section{Introduction}
Semantic segmentation is a fundamental task in computer vision, enabling a pixel-level understanding of visual scenes.
Despite remarkable progress, state-of-the-art methods still depend on large datasets with densely annotated masks, which are expensive and time-consuming to label~\cite{shen2023survey}.
Producing pixel-perfect labels demands substantial manual effort, creating a major bottleneck for deploying segmentation models in new domains.

To reduce this dependence on dense supervision, \emph{Weakly Supervised Semantic Segmentation} (WSSS) has emerged as a practical alternative~\cite{Chen2023WeaklysupervisedSS}.
Instead of relying on full masks, WSSS methods learn from cheaper, indirect cues such as image-level tags~\cite{xu2025toward}, bounding boxes~\cite{lempitsky2010learning, rajchl2016deepcut}, or sparse scribbles~\cite{lin2016scribblesup}.
These weak labels decrease annotation costs but introduce ambiguity about object boundaries and shape, making it difficult for models to produce coherent segmentations without additional structural guidance~\cite{liang2022tree, wu2023sparsely}.

Recently, the landscape of WSSS has been reshaped by foundation models such as the Segment Anything Model (SAM)~\cite{kirillov2023segment}.
Thanks to their prompt-based design and extensive pretraining, foundation models can generate high-quality pseudo-labels from simple user inputs, providing a powerful prior that many WSSS pipelines now exploit. However, these methods suffer from two major limitations. 1) First, they are \emph{tailored to a specific label and cannot handle arbitrary and heterogeneous weak labels}, due to their heuristic prompt-based design. In practice, the availability of labels varies by dataset, so the models should be able to train on all combinations of supervision for best performance.  2) Second, they rely on \emph{frozen foundation models that struggle with domain shifts}, such as medical imaging. 

\begin{figure*}[t]
    \centering
    \resizebox{0.7\textwidth}{!}{
    \input{figures/visualabstract}}
    \caption{Overview of the proposed neurosymbolic weakly supervised segmentation framework. The method operates in two stages: first, the Segment Anything Model (SAM) is fine-tuned using differentiable logic constraints derived from weak annotations and general structural priors to generate refined pseudo-labels. Second, these pseudo-labels are used as ground truth to train a standard segmentation network for final deployment.}
    \label{fig:visabstract}
\end{figure*}
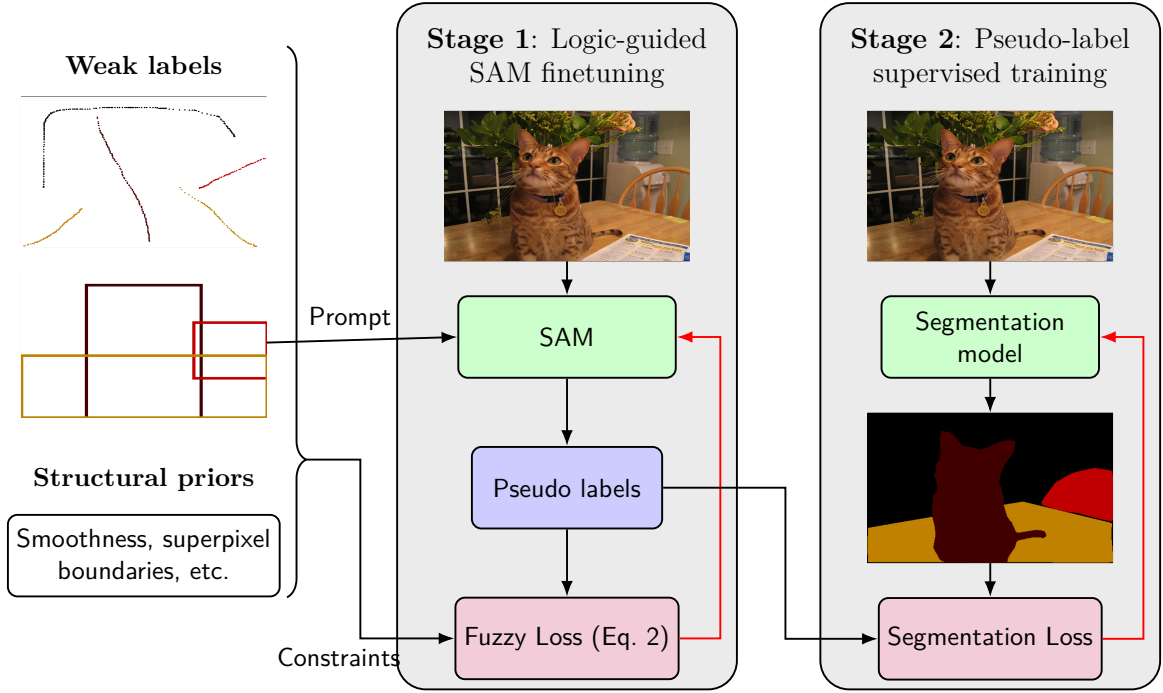

To overcome both limitations, we propose \emph{a neurosymbolic approach for weakly supervised semantic segmentation}. We show that various kinds of weak labels and structural priors can all be expressed in a principled fashion as logical constraints. These constraints are end-to-end differentiable, meaning we can fine-tune the foundation model to respect them. Although inference on logical constraints is NP-hard in general, we leverage fuzzy independence approximations from the neurosymbolic literature~\cite{vankrieken2022analyzing} to keep inference linear-time in the size of the image and the size of the constraint.
%
Although neurosymbolic methods have been applied to the object or scene level ~\cite{manigrasso2021faster, colamonaco2025neurosymbolic}, extending it to the pixel level opens the door to a neurosymbolic segmentation guided by explicit relational rules, a direction that has remained largely unexplored.


Practically, our method follows a two-stage pipeline.
In the first stage, SAM is fine-tuned through differentiable logic losses derived from weak annotations, yielding refined and structurally consistent pseudo-labels.
In the second stage, these pseudo-labels are used as training data for a prompt-free segmentation network, enabling scalable deployment without the need for human prompts.

Experimentally, we show that our neurosymbolic approach to WSSS 1) matches or outperforms the prior WSSS baselines using the same supervision signals, 2) considerably outperforms prior WSSS baselines by using more weak supervision types, where prior methods do not support this, and 3) approaches or even exceeds the performance of fully supervised training, which we hypothesize is partly attributable to imperfections in the available ground-truth annotations.


In summary, the contributions of this work are
\begin{enumerate*}[label=\textit{(\roman*)}]
\item a neurosymbolic formulation of weakly supervised segmentation that fine-tunes SAM with differentiable fuzzy-logic,
\item a unified objective that expresses heterogeneous weak annotations and structural priors as constraints of arbitrary order under a single differentiable relaxation, and
\item empirical evidence that such reasoning-guided learning achieves state-of-the-art performance against other WSSS methods on standard segmentation benchmarks.
\end{enumerate*}

%% file: figures/visualabstract.tex
\begin{tikzpicture}[
    font=\sffamily,
    >=Latex,
    panel/.style={
        draw,
        thick,
        rounded corners=12pt,
        fill=gray!16,
        minimum width=5cm,
        minimum height=10.1cm
    },
    block/.style={
        draw,
        rounded corners=4pt,
        thick,
        align=center,
        minimum width=3.2cm,
        minimum height=1.2cm
    },
    smallblock/.style={
        draw,
        rounded corners=4pt,
        thick,
        align=center,
        minimum width=2.8cm,
        minimum height=1.2cm
    }
]


\node[panel, draw=none, fill=none] (leftpanel) {};
\node[panel, right=1.2cm of leftpanel] (midpanel) {};
\node[panel, right=1.2cm of midpanel] (rightpanel) {};


\node[align=center, font=\large, anchor=north]
    at ($(midpanel.north)+(0,-0.25cm)$)
    {\textbf{Stage 1}: Logic-guided \\ SAM finetuning};

\node[align=center, font=\large, anchor=north]
    at ($(rightpanel.north)+(0,-0.25cm)$)
    {\textbf{Stage 2}: Pseudo-label \\ supervised training};


\node[align=center, font=\bfseries, anchor=north]
    (weaklabelstitle)
    at ($(leftpanel.north)+(0,-0.65cm)$)
    {Weak labels};

\node[draw, inner sep=0pt, anchor=north, below=0.2cm of weaklabelstitle]
      (scribblesimg)
      {\includegraphics[width=3.6cm, height=2.2cm]{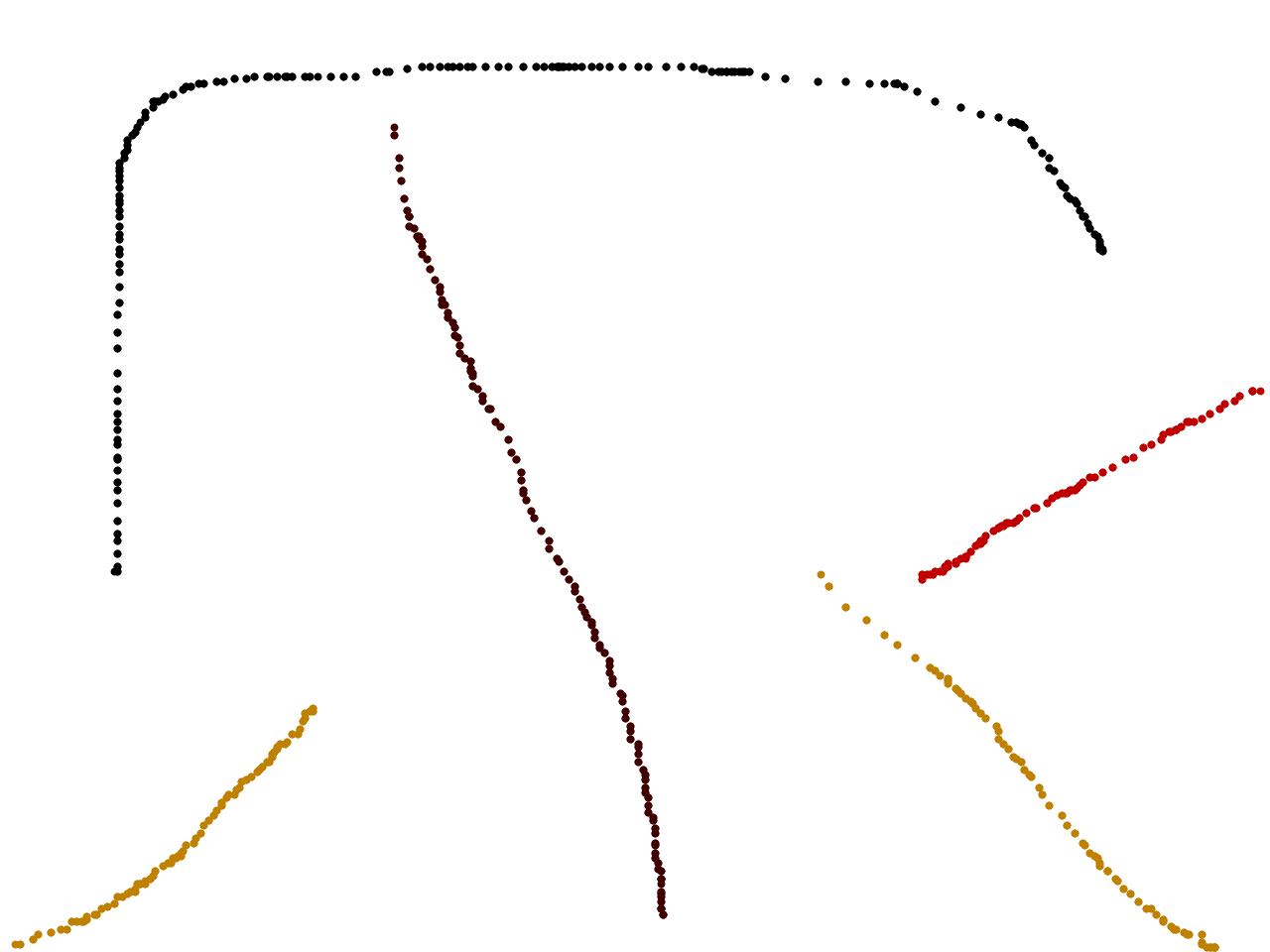}};

\node[draw, inner sep=0pt, below=0.3cm of scribblesimg]
      (bboximg)
      {\includegraphics[width=3.6cm, height=2.2cm]{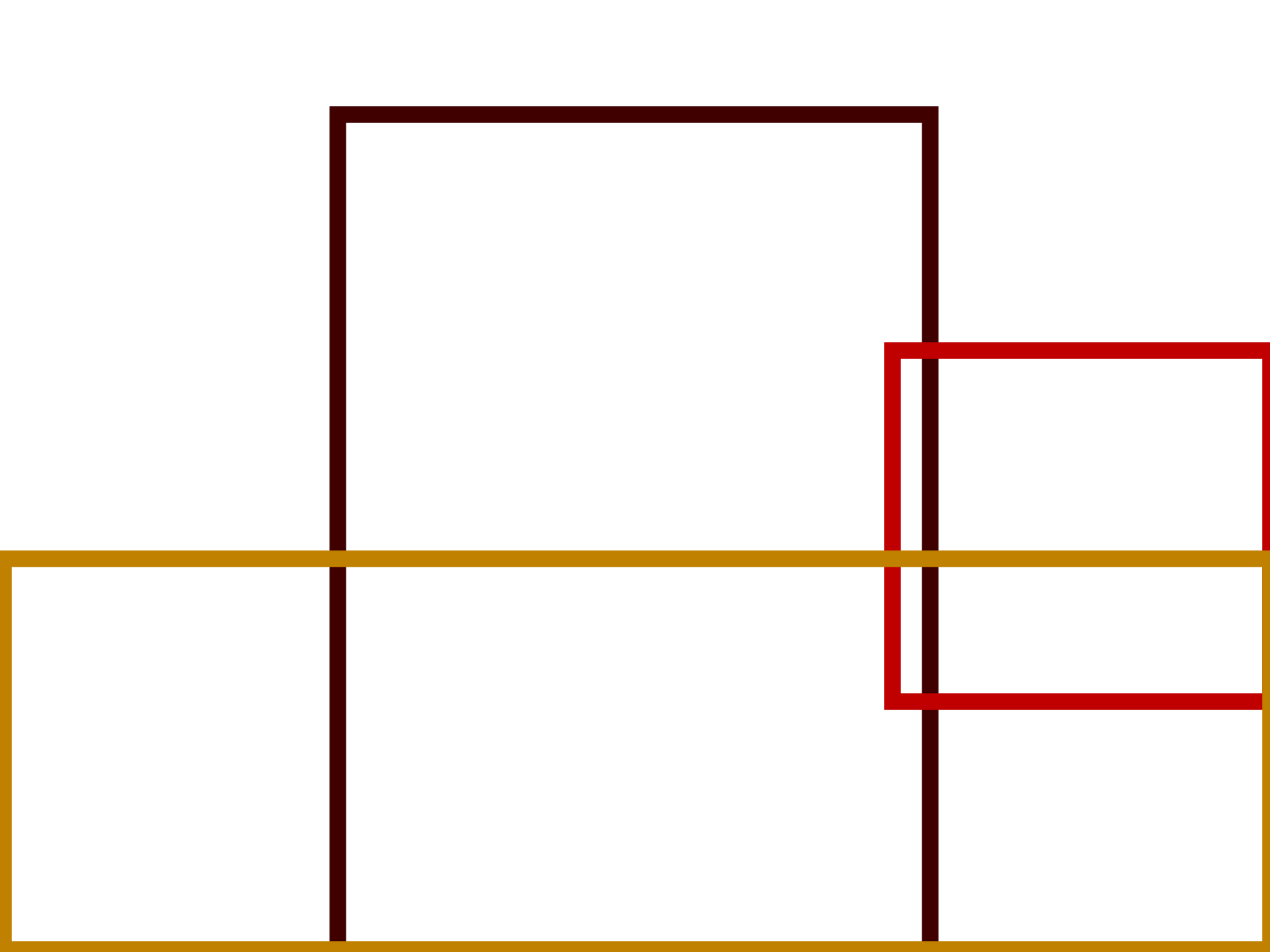}};

\node[align=center, font=\bfseries, below=0.6cm of bboximg]
    (structtitle)
    {Structural priors};

\node[block, fill=white, below=0.2cm of structtitle]
    (logicbox)
    {Smoothness, superpixel \\ boundaries, etc.};


\node[draw, inner sep=0pt, anchor=north]
    (midinputimg)
    at ($(midpanel.north)+(0,-1.6cm)$)
    {\includegraphics[width=3.6cm, height=2.2cm]{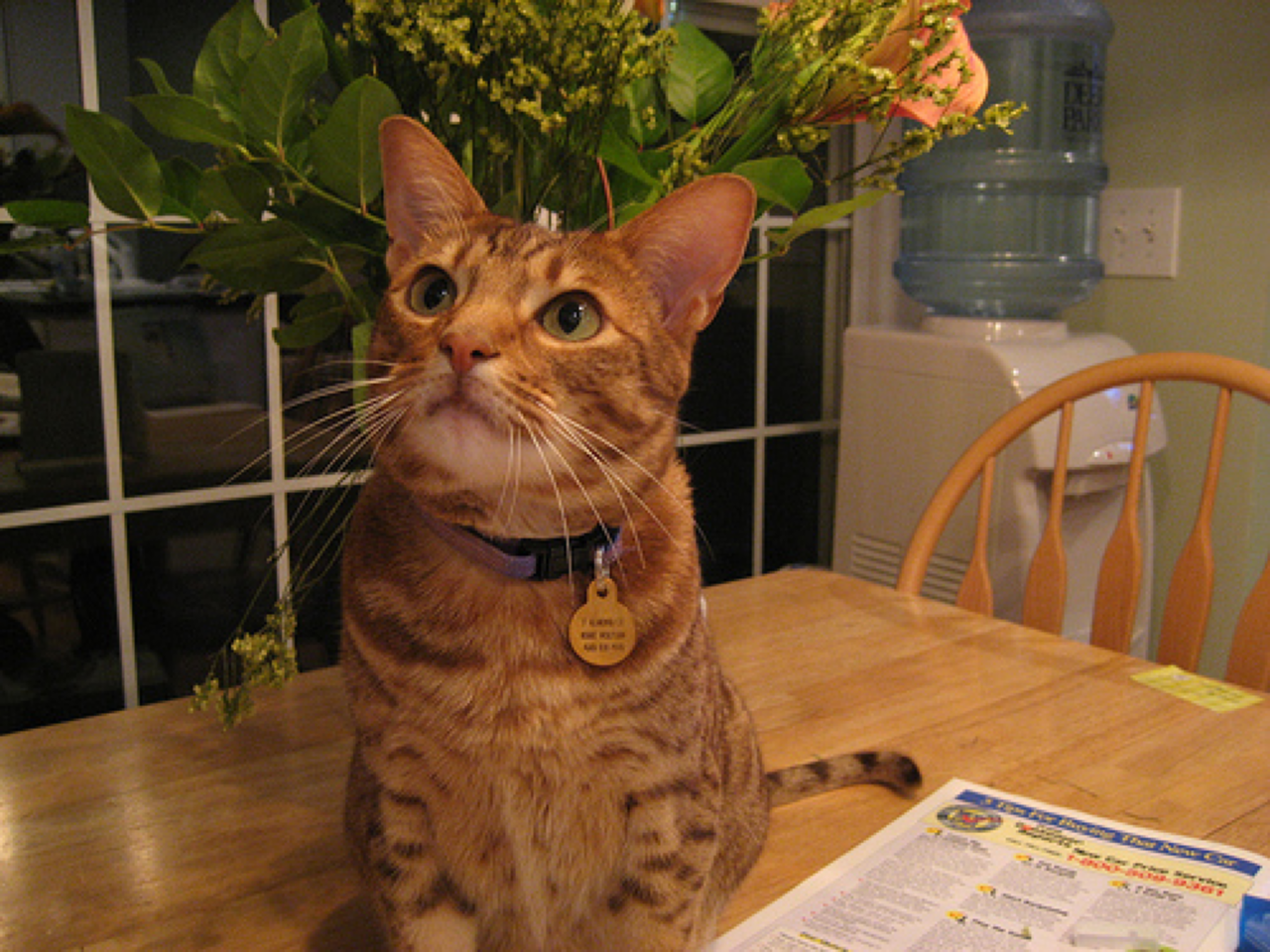}};

\node[block, fill=green!20, below=0.5cm of midinputimg]
    (sam)
    {SAM};

\node[smallblock, fill=blue!20, below=0.99cm of sam]
    (pseudo1)
    {Pseudo labels};

\node[block, fill=purple!20, below=0.99cm of pseudo1]
    (fuzzy)
    {Fuzzy Loss (Eq.~\ref{eq:semantic-loss})};


\node[draw, inner sep=0pt, anchor=north]
    (inputimg)
    at ($(rightpanel.north)+(0,-1.6cm)$)
    {\includegraphics[width=3.6cm, height=2.2cm]{figures/plot_image}};

\node[block, fill=green!20, below=0.5cm of inputimg]
    (segmodel)
    {Segmentation \\ model};

\node[draw, inner sep=0pt, below=0.5cm of segmodel]
      (output)
      {\includegraphics[width=3.6cm, height=2.2cm]{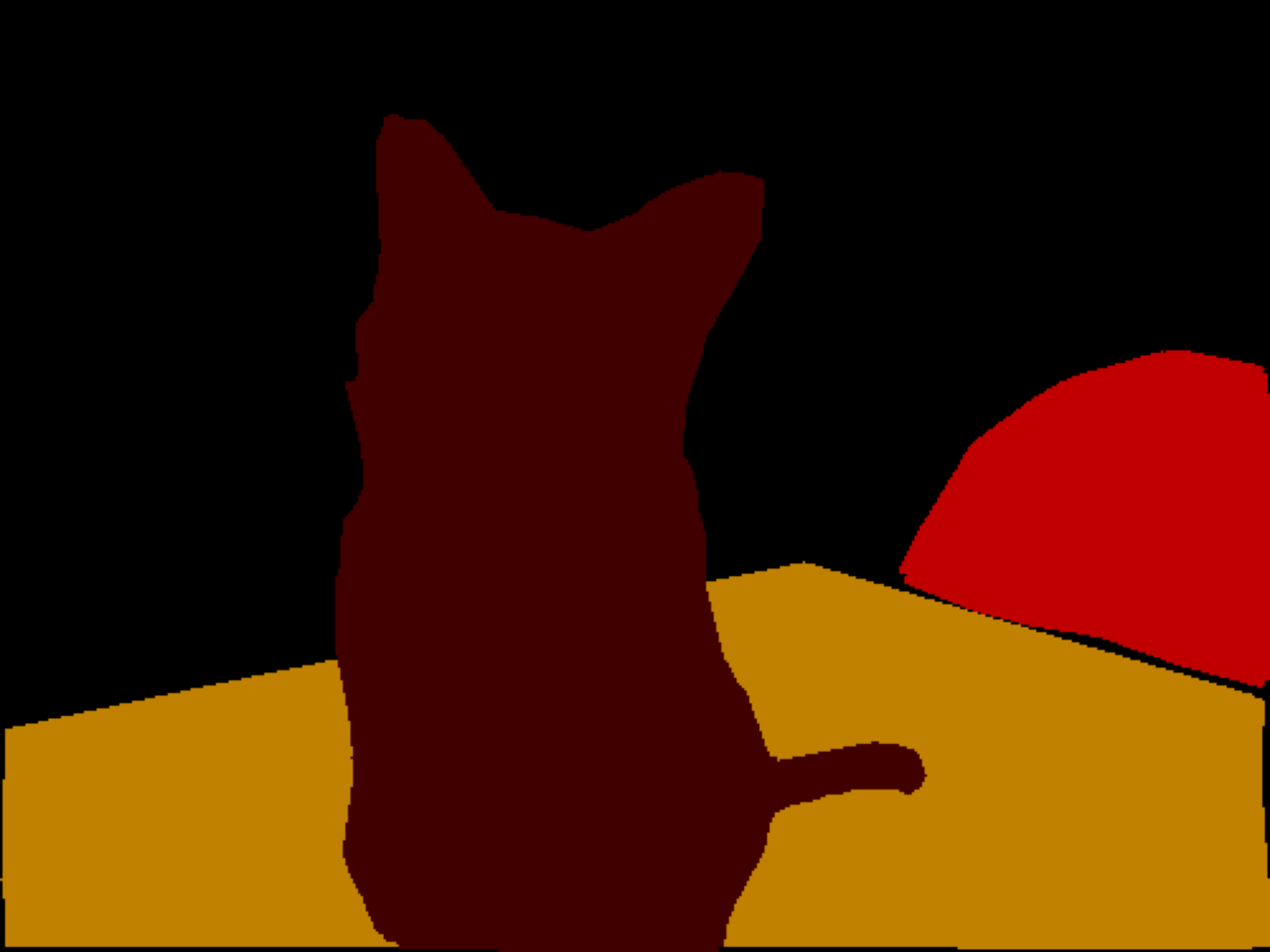}};

\node[block, fill=purple!20, below=0.5cm of output]
    (segloss)
    {Segmentation Loss};


\draw[->, thick]
    (bboximg.east) -- node[above, xshift=-5pt]{Prompt} (sam.west);

\coordinate (btop) at ($(weaklabelstitle.north -| bboximg.east)+(0.3,0)$);
\coordinate (bbot) at ($(logicbox.south -| bboximg.east)+(0.3,0)$);
\draw[thick, decorate, decoration={brace, mirror, amplitude=8pt, aspect=0.25}]
    (bbot) -- (btop);
\draw[->, thick]
    ($(bbot)!0.25!(btop)+(0.28,0)$) -- ++(0.8,0) |- node[below, xshift=-9pt]{Constraints} (fuzzy.west);

\draw[->, thick] (midinputimg) -- (sam);
\draw[->, thick] (sam) -- (pseudo1);
\draw[->, thick] (pseudo1) -- (fuzzy);

\draw[->, thick, red]
    (fuzzy.east) -- ++(0.6,0)
    |- (sam.east);

\draw[->, thick]
    (pseudo1.east) -- ++(1.8,0)
    |- (segloss.west);

\draw[->, thick] (inputimg) -- (segmodel);
\draw[->, thick] (segmodel) -- (output);
\draw[->, thick] (output) -- (segloss);

\draw[->, thick, red]
    (segloss.east) -- ++(0.6,0)
    |- (segmodel.east);

\end{tikzpicture}

%% file: sec/2_related.tex
\section{Related Work}
\noindent\textbf{Weak supervision.} WSSS has been studied extensively with different annotation types. Early work explored bounding boxes~\cite{lempitsky2010learning, rajchl2016deepcut, tian2021boxinst, ji2021weakly}, scribbles~\cite{lin2016scribblesup}, or point-level cues~\cite{wu2023sparsely} as alternatives to dense masks. 
Moreover, many methods exploit class activation maps (CAMs) for image-level labels~\cite{chen2022c} or to propagate labels from sparse scribbles~\cite{ouassit2022brief}. 
For instance, TEL \cite{liang2022tree} optimizes a tree energy loss for scribble supervision, while AGMM~\cite{wu2023sparsely} employs adaptive Gaussian mixtures. 
Concurrently, the field has seen a surge in methods leveraging vision-language models such as CLIP to extract dense pixel-level affinities from simple Image-Level Tags (ILT), as demonstrated by CLIP-ES~\cite{lin2023clip} and CLIP-CPAL ~\cite{tang2024hunting}. However, a primary limitation is that these methods target a single supervision type (\eg, only image-level tags or only scribbles), making them inflexible when diverse annotations are available.
To integrate multiple forms of weak supervision, Xu \etal~\cite{xu2015learning} proposed a unified framework that leverages multiple weak cues 
as linear constraints in a max-margin clustering objective over superpixels. While effective, this discrete formulation reasons over fixed regions rather than individual pixels, and so cannot represent the fine-grained shape and boundary priors central to our approach. We retain the idea of unifying heterogeneous weak cues, but replace the discrete constrained assignment with differentiable fuzzy-logic formulas, a far more expressive constraint language that allows us to fine-tune a deep foundation model.

\paragraph{Foundation models.} The recent shift of task-specific models to foundation models has also been adopted in image segmentation. A notable example is the Segment Anything Model (SAM)~\cite{kirillov2023segment}, which was trained on a billion-mask dataset and generates accurate segmentation masks from simple prompts (points or boxes). Jiang and Yang~\cite{jiang2023segment} showed that prompting SAM with weak labels yields high-quality pseudo-labels that can be used to train a segmentation network in a weakly supervised fashion. 

Although promptable foundation models often outperform traditional pipelines, reliance on a fixed pre-trained model can limit adaptability to out-of-distribution domains~\cite{mazurowski2023segment}. 
Because of this, several works have fine-tuned SAM using large-scale, densely annotated datasets. For instance, MedSAM~\cite{ma2024segment} 
adapt SAM to a wide variety of medical modalities, while methods like SAMed~\cite{zhang2023customized,cheng2023sam} employ low-rank adaptation (LoRA) to efficiently fine-tune the image encoder for medical tasks. However, these specialized models still require pixel-perfect ground truth for fine-tuning. In contrast, our framework adapts foundation models to out-of-distribution domains using only weak supervision.

\paragraph{Neurosymbolic Learning.} In addition to learning from data, neurosymbolic methods integrate prior knowledge and constraints directly into learning. 
For instance, semantic-based regularization~\cite{diligenti2017semantic, diligenti2016learning} and the semantic loss~\cite{xu2018semantic} incorporate logical rules as soft losses. Probabilistic logic programming languages such as DeepProbLog~\cite{manhaeve2018deepproblog} allow rich symbolic priors but also face tractability issues for high-dimensional problems such as segmentation~\cite{maene2024hardness}, whereas differentiable fuzzy logic remains tractable in the propositional case~\cite{fischer2019dl2, badreddine2022logic, vankrieken2022analyzing}. 

Building on these tractable frameworks, some neurosymbolic methods have been adapted for vision tasks. In classification, Shukla \etal~\cite{shukla2023unified} introduced a unified differentiable loss that handles multiple types of weak labels, while FasterLTN~\cite{manigrasso2021faster} integrates a Faster R-CNN backbone with Logic Tensor Networks to enable visual learning under first-order constraints. Such approaches have seen limited use in segmentation, given the complexity of pixel-level reasoning. Concurrently, Bergamin \etal~\cite{bergamin2025integrating} explored medical semantic segmentation using Logic Tensor Networks. While promising, their approach operates in a fully supervised setting and does not tackle weakly supervised learning.

Enforcing spatial coherence traditionally relies on post-processing steps such as CRFs~\cite{krahenbuhl2011efficient} or affinity matrices~\cite{ahn2018learning}. 
More recently, several works instead embed weak labels and structural priors directly into the training loss, either as relaxed MRF/CRF regularizers~\cite{tang2018regularized} or as inequality constraints on output statistics~\cite{kervadec2019constrained,kervadec2020bounding}. We generalize this regularized-loss paradigm: we express unary, pairwise, and higher-order constraints including domain-specific priors, which standard pairwise potentials do not express and which otherwise require special-purpose higher-order constructions.


%% file: sec/3_preliminaries.tex
\section{Preliminaries}

\paragraph{Notation.} We write an integer interval as $[a..b] = \{a, a+1, \dots,  b\}$. We write a random variable $\Xvar$ in uppercase and a value $\xvar \in \dom{\Xvar}$ in lowercase, where $\dom{\Xvar}$ is the sample space of $\Xvar$. Sets of random variables $\Xvars$ and sets of values $\xvars$ are denoted in bold.


\paragraph{Image Segmentation.} In the image segmentation setting, we have a set of random variables $\Xvars$ for the pixel values and a set of random variables $\Yvars$ for the pixel labels, such that $\xvars \in \dom{\Xvars}$ is an image and $\yvars \in \dom{\Yvars}$ is a segmentation map. Specifically, $\Xvar_{i,j} \in \Xvars$ corresponds to the pixel vector at spatial coordinates $(i,j)$ in the $n \times m$ grid with $c$ channels ($c=3$ for RGB). Consequently, the pixel sample space is $\dom{\Xvar_{i,j}} = [0, 1]^c$. 
The label sample space $\dom{\Yvar_{i,j}}$ is finite and depends on the dataset. For example, $\dom{\Yvar_{i,j}} = \{ \text{background}, \text{cat}, \text{dog}\} $.

We are interested in the posterior distribution $p(\Yvars \mid \xvars)$, which conditions the distribution over segmentation maps on the input image $\xvars$. To this end, we assume a differentiable function $f_\theta$ with weights $\theta$, where given an image $\xvars \in \dom{\Xvars}$, $p(\Yvars \mid \xvars) = f_\theta(\xvars)$ is the desired distribution.


\paragraph{Propositional Logic.} 
A propositional formula $\phi$ is inductively defined as either an atom, a negation $\neg \phi$, a conjunction $\phi_1 \land \phi_2$, a disjunction $\phi_1 \lor \phi_2$, or an implication $\phi_1 \Rightarrow \phi_2$. An atom either states that a specific pixel has a certain label $\Yvar = \yvar$, or that two pixels have the same label $\Yvar_1 = \Yvar_2$.  A formula $\phi$ is satisfied in a segmentation map $\yvars$, written $\yvars \models \phi$, if the formula $\phi$ evaluates to true under the usual semantics. A formula $\phi_1$ entails another formula $\phi_2$, written $\phi_1 \models \phi_2$, if $\yvars \models (\phi_1 \Rightarrow \phi_2)$ for any segmentation map $\yvars$.

\begin{example}
    Consider a segmentation map $\yvars$ specified over the domain $\Omega(Y_{i,j}) = \{ \text{background}, \text{cat}, \text{dog} \}$.
    We create a formula $\phi_\mathbf{cat}$ to check if the entire diagonal of the image is all classified as cat. \[\phi_\mathbf{cat} \coloneq 
    \bigwedge_{i=1}^n (Y_{i,i} = \text{cat})\] 
    On a tiny $2 \times 2$ image specified as $\yvars_{1,1} = \text{cat}$, $\yvars_{1,2}=\text{background}$, $\yvars_{2,1} = \text{background}$, $\yvars_{2,2}=\text{cat}$, it holds that $\yvars \models \phi_\mathbf{cat}$.
\end{example}

\paragraph{Semantic loss.} Instead of a fixed segmentation map $\yvars$, we have a distribution over segmentation maps $\Yvars$. So, the probability of $\phi$ being satisfied becomes the expectation $p(\phi) = \mathbb{E}_{\yvars \sim \Yvars}[\yvars \models \phi]$. The probability of a formula $\phi$ being satisfied conditioned on an input image $\xvars$ is then

\begin{align}
\label{eq:sem-loss1}
\begin{split}
p(\phi \mid \xvars) &= \sum_{\yvars \in \dom{\Yvars}} p(\phi \mid \yvars)\, p(\yvars \mid \xvars) \\
&= \mathbb{E}_{\yvars \sim f_\theta(\xvars)} [ \yvars \models \phi ].
\end{split}
\end{align}

In the second equality, we use the fact that the formula $\phi$ and image $\xvars$ are conditionally independent given the segmentation map $\yvars$.
During training, we simply optimize the negative log-likelihood of the formula being satisfied given the input image. This is also known as the semantic loss~\cite{xu2018semantic}.
\begin{equation}\label{eq:semantic-loss}
\mathcal{L}(\xvars, \phi) \coloneq - \log p(\phi \mid \xvars)
\end{equation}

In our formulation above, all logical formulas are assumed correct and we hence do not weigh them in the loss objective. However, in practice, weak annotations and structural priors may be imperfect. In \cref{subsec:inexact-constraints} we provide a detailed theoretical discussion on how to handle inexact constraints, including potential loss weighting strategies based on their correlation with the ground truth.

\paragraph{Approximation.} Computing $p(\phi \mid \xvars)$ exactly is \#P-complete~\cite{valiant1979complexity} and it is hence often necessary to approximate. For this reason, we introduce conditional independence assumptions, see \cref{app:CIA}. These independence assumptions can also be understood as fuzzy logic semantics, specifically the product t-norm~\cite{novak1999mathematical}. In other words, the approximation becomes more faithful as the predictions of the neural network become more confident (\ie, probabilities closer to 0 or 1). Optimizing the semantic loss using fuzzy semantic is known as \emph{semantic-based regularization}~\cite{diligenti2017semantic}.

%% file: sec/4_constraints.tex
\section{Pixel-level Logical Constraints}\label{sec:pixel-constraints}

In this section, we discuss how to express various types of segmentation labels as logic formulas. These formulas can then be optimized with the semantic loss to perform weakly supervised learning.
Full supervision is a special case of our logical weak supervision. In that case, we get the ground truth labels $\yvars^*$ for all pixels. So, we get the logical formula 
\begin{equation}
\phi_\textbf{fs} \coloneq \bigwedge_{\substack{i \in [1..n] \\ j \in [1..m]}} (\Yvar_{i,j} = \yvar^*_{i,j}).    
\end{equation}
The indices $i$ and $j$ range over all rows and columns of the image, respectively. The loss $\mathcal{L}(\xvars , \phi_\textbf{fs})$ equals the usual cross-entropy loss (\cf \cref{app:equivalence}).

\subsection{Weakly Supervised Constraints} \label{sec:WSC}

\paragraph{Scribbles.} A scribble $s$ is a subset of pixel locations that should have a target class $y^*$. So the scribble constraint is similar to full supervision, except that the conjunction does not range over all pixels but only those in the scribble $s$, leaving unmarked pixels unconstrained.
\begin{equation}
 \phi_\textbf{scribbles} \coloneq \bigwedge_{(i,j) \in s} (\Yvar_{i,j} = y^*)      
\end{equation}


\paragraph{Bounding boxes.} A bounding box is a rectangle, defined by two corners $(i_1, j_1)$ and $(i_2, j_2)$, and a class $\yvar^*$. The simplest constraint would be to assume that at least one of the pixels in the bounding box has the target class $\yvar^*$.
\begin{equation}
\phi_\textbf{bbox\,shallow} \coloneq \bigvee_{\substack{i \in [i_1..i_2] \\ j \in [j_1..j_2]}} (\Yvar_{i,j} = \yvar^*)
\end{equation}

The above constraint is rather lenient, and it is unlikely that only a single pixel in a full bounding box has the target label. A more stringent assumption ($\phi_\textbf{bbox}$) is that the bounding box is tight. In that case, each row and each column of the bounding box contain a pixel with the target class $\yvar^*$. It follows that $\phi_\textbf{bbox} \models \phi_\textbf{bbox\,shallow}$.

Lastly, given a set of bounding boxes, the background constraint $\phi_\textbf{background}$ states that all the area outside all bounding boxes has the background class as the target label, which in essence is the same as a large scribble constraint.

\subsection{Unsupervised Constraints} \label{sec:UC}
Beyond the supervision from weak annotations, segmentation masks typically satisfy structural properties. We incorporate these priors as unsupervised logical constraints (\ie, formulas that do not rely on any label) to regularize the model's predictions and propagate information across pixels. Additional formal definitions are detailed in \cref{app:method_formulation}. 

\paragraph{Smoothness.} Segmentation maps typically satisfy some smoothness constraints. One example is the region filling constraint ($\phi_\textbf{fill}$), which enforces that if all neighbors of a pixel share the same class, the central pixel must also belong to that class, thereby closing small holes in the segmentation mask. To impose stricter consistency and prevent isolated artifacts, we introduce a stronger neighborhood constraint ($\phi_\textbf{neighborhood}$), requiring that for each pixel, at least one neighbor should have the same label.

 \paragraph{Superpixels.} Superpixels~\cite{ren2003learning} are a classic computer vision technique to group perceptually similar pixels. We leverage superpixels to introduce a border-preserving constraint ($\phi_\textbf{borders}$), which encourages class transitions to occur only at superpixel boundaries by penalizing adjacent pixels within the same superpixel that are assigned different labels.

 \paragraph{Corners.} Our framework's flexibility makes it trivial to integrate domain-specific geometric priors when such prior knowledge is available. For instance, in medical datasets such as REFUGE2, the anatomical structures of interest (\eg, optic discs and cups) are known to be roughly circular or elliptical. To exploit this, we introduce a corner constraint ($\phi_\textbf{corners}$). Because a circular object rarely occupies a bounding box's extreme corners, we mathematically define an inscribed ellipse within the box. The constraint enforces that pixels inside the box but outside this ellipse must \textit{not} belong to the target class.

%% file: sec/5_logic-guided.tex
\section{Logic-Guided Pseudo-Label Refinement}

\begin{figure*}
    \centering
    \includegraphics[width=0.80\linewidth]{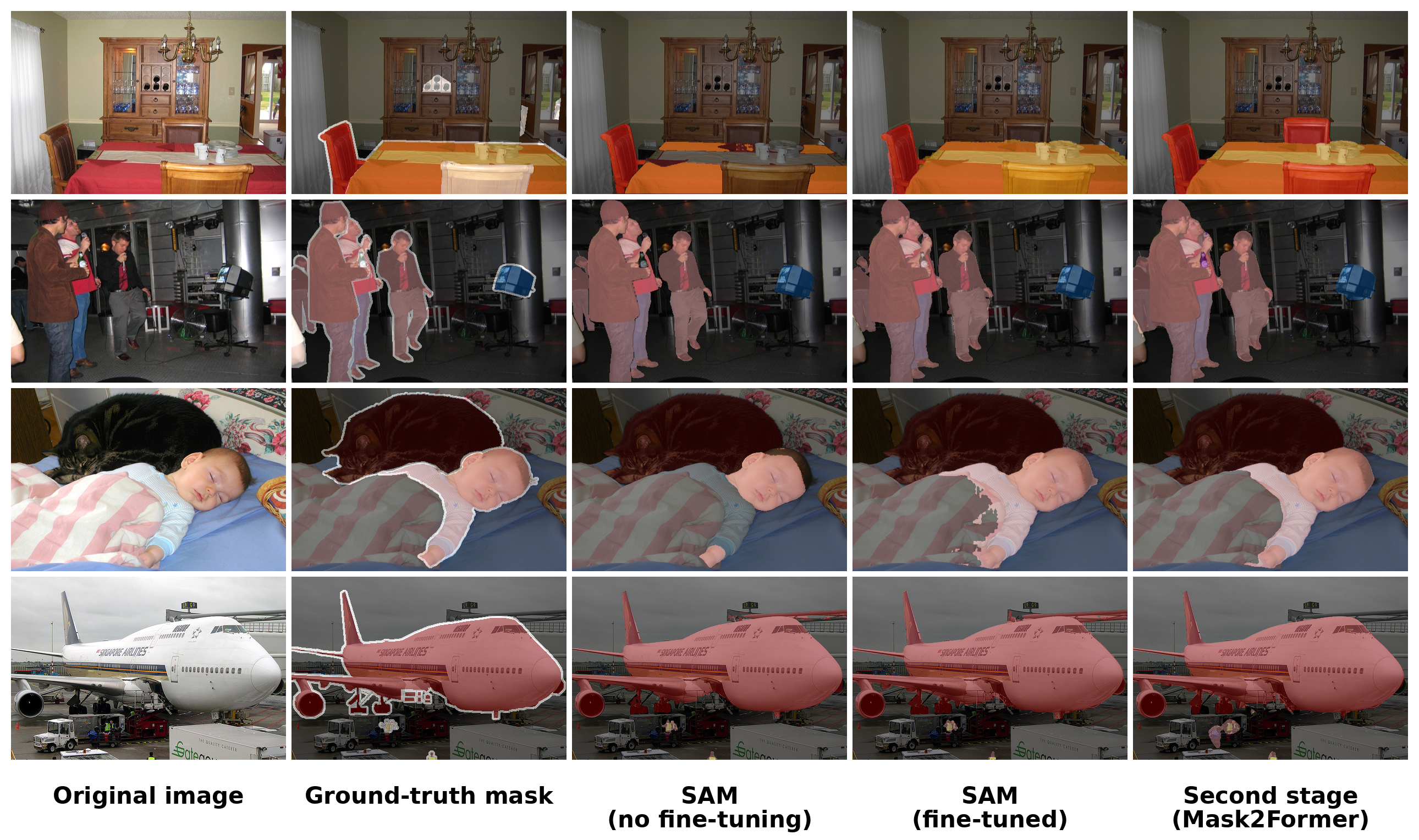}
    \caption{Qualitative comparison of our two-stage weakly supervised segmentation pipeline on the Pascal VOC 2012 validation set. SAM is prompted with bounding boxes.
    }
    \label{fig:examples_pipeline}
\end{figure*}

We now propose the logic-guided weakly supervised segmentation framework that integrates the differentiable constraints from \cref{sec:WSC,sec:UC} with the Segment Anything Model (SAM).
The method turns weak supervision signals (\ie, bounding boxes, scribbles) into logical constraints that are optimized jointly with SAM’s decoder.
The training proceeds in two stages.
\begin{enumerate*}[label=(\roman*)]
    \item \emph{Logic-guided fine-tuning}: SAM receives bounding-box prompts and is optimized to satisfy a set of fuzzy-logic constraints that encode the available weak labels and priors.
    \item \emph{Pseudo-label supervised training}: the fine-tuned SAM generates pseudo-labels, which are used to train a standard segmentation network.
\end{enumerate*}

\subsection{Logic-Guided SAM Fine-tuning}
In the first stage, SAM is adapted to the weakly supervised setting by optimizing its outputs with respect to a set of differentiable logical constraints. Given a training set $\mathcal{D}=\{(\xvars_i, \phi_i)\}^n_{i=1}$ of images and formulas, fine-tuning aims to learn a function $f_\theta$ that minimizes the fuzzy constraint satisfaction objective.
\begin{equation}
    \label{eq:constraints_comb}
    \theta^* = \underset{\theta}{\operatorname{arg\,min}} \sum_{i=1}^{n} \mathcal{L}(\xvars_i, \phi_i; \theta) 
\end{equation}

Note that the constraints of each image may differ, depending on the available weak annotations, and that each $\phi_i$ may include multiple constraints conjoined together. For example, to use both scribbles and bounding boxes, we may simply set $\phi = \phi_\textbf{scribbles} \land \phi_\textbf{bbox}$.
Rather than evaluating these formulas as strict True/False statements, they are relaxed into continuous functions using product t-norm semantics. Simply put, this replaces the discrete logical operations with differentiable operations (\eg, a logical "AND" becomes a multiplication). This continuous formulation allows the entire objective to act as a loss function that can be optimized end-to-end using standard gradient-based optimizers. Additional details of the differentiable implementation are provided in \cref{app:diff_imp}.

During fine-tuning, SAM receives prompts derived directly from the weak annotations (\ie, bounding boxes). To lower the computational overhead, only the mask decoder is fine-tuned, while the image encoder remains frozen.

\subsection{Supervised Segmentation from Pseudo-labels}
The second stage follows a conventional fully supervised training pipeline. The fine-tuned SAM from the first stage generates pseudo-labels for the entire training set. These pseudo-labels are then treated as the ground-truth labels. A standard segmentation architecture, such as Mask2Former or DeepLabV2, is trained from scratch on this dataset of images and their corresponding pseudo-labels. The model is trained using the conventional cross-entropy loss, learning to replicate the high-quality predictions from the logic-guided SAM. The resulting model is the final output of our framework: an efficient, prompt-free network that benefits from both the foundation model as well as the logic-guided fine-tuning.


%% file: sec/6_experiments.tex
\section{Experiments}
We design our experiments for the following four primary research questions.
\begin{enumerate}[label={[\textbf{RQ\arabic*}]}, leftmargin=*, align=left]
    \item Does logic-guided fine-tuning produce pseudo-labels closer to the ground truth than prompt-based methods without domain-specific fine-tuning?
    \item Can logic-guided fine-tuning with weak labels achieve competitive performance against medical SAM variants fine-tuned with dense annotations?
    \item Does training a segmentation network on our pseudo-labels improve over existing WSSS and fully supervised methods? 
    \item What is the specific contribution 
    of each individual logical constraint on the pseudo-label performance?
\end{enumerate}

\begin{figure*}
    \centering
    \includegraphics[width=0.8\linewidth]{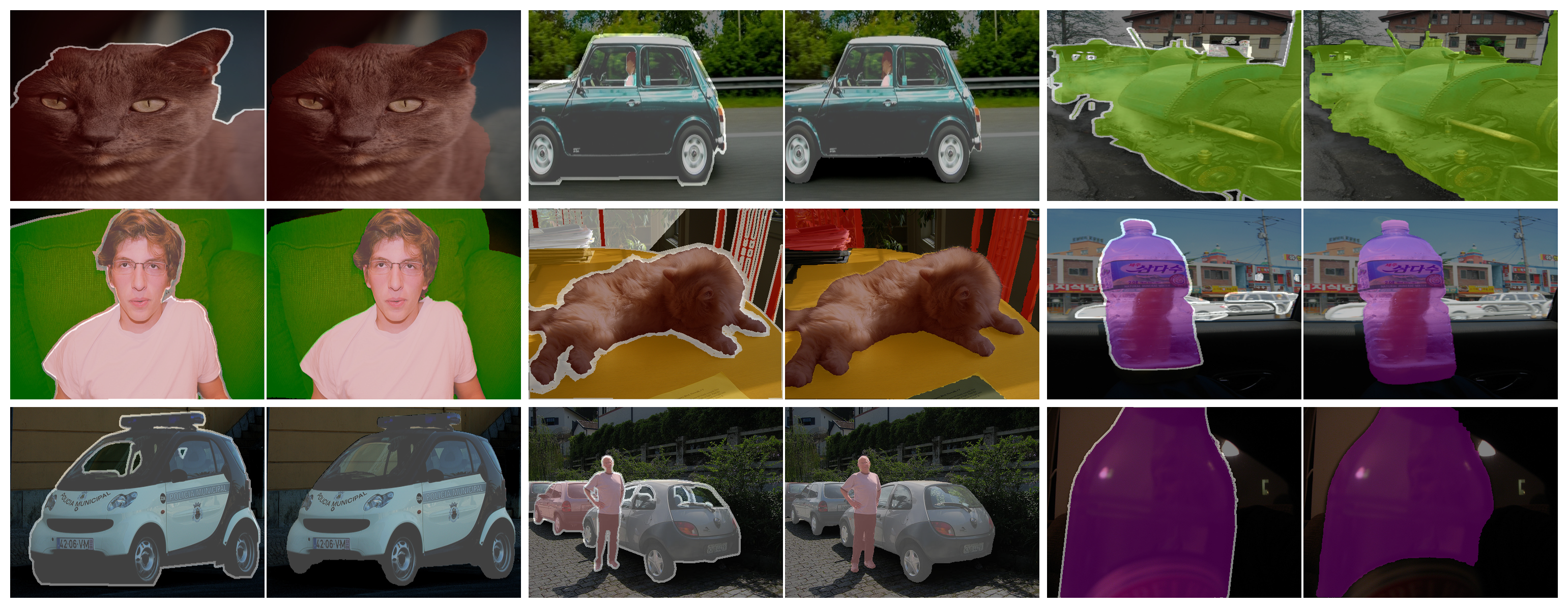}
    \caption{Qualitative comparison of selected samples from the Pascal VOC 2012 training set. The columns alternate between the original human-annotated ground truth and the pseudo-labels produced by our logic-guided fine-tuned SAM. Interestingly, in several instances like the ones shown here, the pseudo-labels generated by our model seem to provide a better segmentation than the original annotation, capturing thin structures and boundaries that the ground truth omits or outlines coarsely.}
    \label{fig:good_pseudolabels}
\end{figure*}


\paragraph{Datasets.} We evaluate our framework on two benchmarks: Pascal VOC 2012~\cite{everingham2010pascal} for natural images and the REFUGE2 retinal dataset~\cite{fang2022refuge2} to test adaptability in a specialized medical domain. For Pascal VOC, following standard WSSS approaches~\cite{ji2021weakly, lin2025semantic}, we train on the widely used augmented dataset, reserving 10\% of the images for validation. We utilize the available ground-truth bounding boxes alongside scribbles from prior work~\cite{lin2016scribblesup}. In contrast, REFUGE2 provides only dense masks, so we synthetically derive bounding box and point-level supervision directly from the ground truth (see~\cref{app:training_details} for details). At test time, to simulate real-world annotation noise, we evaluate our first-stage model's robustness on REFUGE2 using non-tight bounding boxes (75\% overlap). This aligns with recent robustness evaluations of medical foundation models~\cite{wu2025medical}.

\paragraph{Baselines and Setup.} For our first-stage foundation model, we utilize the standard SAM for the Pascal VOC dataset. To better handle the domain shift of medical imaging, we adopt the pre-trained MedSAM as our base architecture for the REFUGE2 dataset. During fine-tuning, our Pascal VOC setup relies on standard structural and weak label constraints ($\phi_\textbf{neighborhood}$, $\phi_\textbf{bbox}$, $\phi_\textbf{fill}$, $\phi_\textbf{background}$, $\phi_\textbf{scribbles}$, $\phi_\textbf{borders}$), while our REFUGE2 experiments additionally incorporate the domain-specific $\phi_\textbf{corners}$ prior. We benchmark our approach against a diverse set of baselines: on Pascal VOC, we compare against SAM pipelines~\cite{jiang2023segment} (including different prompting techniques and cross-entropy based finetuning) and established WSSS methods; on REFUGE2, we evaluate against zero-shot MedSAM, fully supervised MedSAM, and other medical foundation models. Finally, for our prompt-free second-stage network, we train DeepLabV2~\cite{chen2017deeplab} and Mask2Former~\cite{cheng2022masked}, standard architectures that ensure fair comparisons with published results.

\paragraph{Evaluation.} Following standard semantic segmentation practice, we report the mean Intersection over Union (mIoU). IoU measures the intersection-to-union ratio of predicted and ground-truth regions. Additionally, for the medical imaging benchmark, we evaluate our performance using the Dice score, which measures spatial overlap and serves as a standard evaluation criterion in the medical domain. To answer our final research question, we conduct an ablation study measuring the impact of different constraints. Since Pascal VOC 2012 test set ground-truth annotations are unknown, we obtained the results from the official evaluation server.

%% file: sec/7_results.tex
\input{tables/first_stage_pascal}

\input{tables/second_stage_pascal}

\section{Results}
We organize our results around the four research questions, reporting pseudo-label quality and final segmentation accuracy on Pascal VOC and REFUGE2. Additional results are available in \cref{app:additional_res}.

\paragraph{[RQ1] Logic-guided fine-tuning significantly improves pseudo-label quality. } As shown in \cref{tab:first_stage_pascal}, our logic-guided pseudo-labels on Pascal VOC achieve 94.5\% mIoU. This represents a substantial improvement over purely prompt-based methods, surpassing the 91.8\% mIoU achieved by the baseline SAM when prompted with bounding boxes and scribbles. This confirms that integrating multiple weak labels through differentiable logical constraints produces pseudo-labels that are much closer to the ground truth than relying solely on the foundation model's zero-shot priors. Additionally, as shown in \cref{fig:good_pseudolabels}, there are cases where our generated pseudo-labels can even surpass the dataset's ground truth. 

\paragraph{[RQ2] Logic-guided fine-tuning achieves competitive performance against fully supervised medical SAM models.} 
As detailed in \cref{tab:first_stage_refuge}, fine-tuning MedSAM with our weakly supervised logic constraints reaches 87.5\% mean Dice and 77.9\% mIoU on REFUGE2. Under identical weak supervision, this outperforms a MedSAM baseline fine-tuned with partial cross-entropy (78.3\% mean Dice), isolating the benefit of our logical constraints over naive partial supervision. The gain over zero-shot prompting is larger still, as directly prompting MedSAM with bounding boxes yields only 47.7\% mean Dice.
Remarkably, this performance closely approaches the upper bound of the same MedSAM architecture trained with full pixel-wise annotations (88.7\% mean Dice). Furthermore, our weakly supervised model outperforms other densely supervised variants such as SAMed (87.1\% mean Dice), while remaining highly competitive with other state-of-the-art models like SAM-Med2D. 

\input{tables/first_stage_refuge}

\input{tables/ablations}

\paragraph{[RQ3] Models trained on our pseudo-labels establish a new state-of-the-art for WSSS.}
On Pascal VOC (\cref{tab:second_stage_pascal}), our framework establishes highly competitive results across both validation and test sets. With a DeepLabV2 architecture, our second-stage model reaches 79.7\% mIoU on the validation set, exceeding the fully supervised version of the same architecture (77.7\%). On the test set, this configuration achieves a strong 78.5\% mIoU, nearly reaching the fully supervised upper bound of 79.7\%. Furthermore, it demonstrates clear superiority over all other weakly supervised baselines sharing the DeepLabV2 architecture. The improvements are even more pronounced when upgrading to the modern Mask2Former architecture: our prompt-free model achieves an impressive 88.6\% and 89.1\% mIoU on the validation and test sets, respectively. This not only significantly outperforms recent WSSS approaches, but remarkably surpasses the densely supervised Mask2Former baseline across both splits.
On the REFUGE2 dataset (\cref{tab:second_stage_refuge} in Appendix), our second-stage network trained solely on our pseudo-labels achieves a mean Dice of 87.1\%. While a gap remains between our approach and fully supervised models like TransUNet (90.3\%), it is highly competitive and slightly outperforms the fully supervised ResUNet baseline (86.5\%). 

\paragraph{[RQ4] The border and scribble constraints provide the largest impact.} The ablation study in \cref{tab:constraints_ablation} on Pascal VOC pseudo-labels answers our final research question. The full set of constraints ($\mathcal{F}$) achieves 94.50\% mIoU on the train set. The most critical constraints are $\phi_{\textbf{borders}}$ and $\phi_{\textbf{scribbles}}$, as removing them causes performance drops to 90.96\% (-3.54\%) and 91.73\% (-2.77\%), respectively. This indicates that the direct supervision from scribbles and the structural guidance from the border-preserving constraint are key drivers of quality. The bounding box constraint $\phi_{\textbf{bbox}}$ surprisingly does not provide a notable contribution (-0.66\%). This is because the box prompt and $\phi_{\textbf{background}}$ already carry most of the box information. Other structural priors also offer small gains. Furthermore, a supplementary ablation on the REFUGE2 dataset confirms the necessity of our domain-specific prior. Without the $\phi_\textbf{corners}$ constraint, the mIoU on the train set drops from 77.9\% to 32.5\%, highlighting its role.

%% file: tables/first_stage_pascal.tex
\begin{table}[t]
\centering
\caption{Quantitative evaluation of first-stage pseudo-label quality on the Pascal VOC 2012 train set. We report the mIoU of the pseudo-labels (PL) generated by our logic-guided fine-tuned SAM, compared against established weakly supervised baselines.}
\begin{tabular}{l l r}
\toprule
\textbf{Method} & \textbf{Annotations} & \textbf{mIoU\textsubscript{PL}(\%)}\\
\midrule
Full supervision & Pixel-wise & 100.0\\
\midrule
Box2TagBack~\cite{ji2021weakly} & Boxes & 90.2\\
Sun \etal~\cite{sun2023alternative} & ILT & 88.3\\
VPL~\cite{xu2025toward} & ILT, Language & 80.1\\
\midrule
SAM~\cite{kirillov2023segment,jiang2023segment} & ILT & 61.9\\
 & Points & 71.7\\
 & Scribbles & 89.7\\
 & Boxes & 91.5\\
\quad + combined prompt & Boxes, Scribbles & 91.8\\
\quad + CE finetune & Boxes, Scribbles & 88.4\\
\midrule
\textbf{This method} & Boxes, Scribbles & \textbf{94.5}\\
 & Scribbles & 93.4\\
 & Boxes & 91.7\\
\bottomrule
\end{tabular}
\label{tab:first_stage_pascal}
\end{table}

%% file: tables/second_stage_pascal.tex
\begin{table*}[t]
\centering
\caption{Quantitative comparison of final second-stage segmentation results on the Pascal VOC 2012 validation and test set. We report the mIoU of our prompt-free networks trained entirely on our logic-guided pseudo-labels, compared against weakly supervised and fully supervised methods. The specific network architecture used for each method is denoted to ensure fair comparison. The $*$ denotes the use of CRF, while $\dagger$ indicates that the method reports results using DeepLabV3+ for the validation set.}
\label{tab:second_stage_pascal}
\begin{tabular}{l l l r r}
\toprule
\textbf{Method} & \textbf{Annotations} & \textbf{Network} & \textbf{mIoU\textsubscript{val}(\%)} & \textbf{mIoU\textsubscript{test} (\%)} \\
\midrule
TEL~\cite{liang2022tree} & Scribbles & DeepLabV2$^\dagger$ & 77.1\phantom{ / 00.0$^*$} & 74.8 / 76.0$^*$\\
AGMM~\cite{wu2023sparsely} & Scribbles & DeepLabV3+  & 76.4\phantom{ / 00.0$^*$} & \multicolumn{1}{c}{-}\\
Box2TagBack~\cite{ji2021weakly} & Boxes & DeepLabV2 & 75.7\phantom{ / 00.0$^*$} & 75.5\phantom{ / 00.0$^*$} \\
Sun \etal~\cite{sun2023alternative} & ILT & DeepLabV2 & 77.2$^*$ & 77.1$^*$\\
rloss with CRF~\cite{tang2018regularized} & Scribbles & DeepLabV2 & 73.0 / 75.0$^*$ & \multicolumn{1}{c}{-}\\
Jiang \etal
~\cite{jiang2023segment} & ILT & DeepLabV2 & 71.1\phantom{ / 00.0$^*$} & 72.2\phantom{ / 00.0$^*$}\\
\  & Points & DeepLabV2 & 69.0\phantom{ / 00.0$^*$} & 68.7\phantom{ / 00.0$^*$}\\
\  & Scribbles & DeepLabV2 & 75.9\phantom{ / 00.0$^*$} & 76.6\phantom{ / 00.0$^*$}\\
\  & Boxes & DeepLabV2 & 76.3\phantom{ / 00.0$^*$} & 75.8\phantom{ / 00.0$^*$}\\
Combined prompt & Boxes, Scribbles & DeepLabV2 & 76.1 / 76.7$^*$ & \multicolumn{1}{c}{-} \\ 
Full supervision & Pixel-wise & DeepLabV2 & 76.4 / 77.7$^*$ & \textbf{79.7$^*$} \\
\textbf{This method} & Boxes & DeepLabV2 & 76.4 / 77.4$^*$  & \multicolumn{1}{c}{-}\\
\ & Scribbles & DeepLabV2 & 77.4 / 78.5$^*$  & \multicolumn{1}{c}{-}\\
\  & Boxes, Scribbles & DeepLabV2 & \textbf{78.2} / \textbf{79.7$^*$}  & \textbf{77.0} / 78.5$^*$\\
\midrule
VPL~\cite{xu2025toward} & ILT, Language & CLIP & 79.3$^*$ & 79.0$^*$\\ \midrule
FMA-WSSS \cite{yang2024foundation} 
& ILT & Mask2Former & 82.6\phantom{ / 00.0$^*$} & 81.6\phantom{ / 00.0$^*$}\\
DHR \cite{jo2024dhr} 
& ILT & Mask2Former & 82.3\phantom{ / 00.0$^*$} & 82.3\phantom{ / 00.0$^*$}\\
Combined prompt & Boxes, Scribbles & Mask2Former & 85.2\phantom{ / 00.0$^*$} & \multicolumn{1}{c}{-} \\ 
Full supervision & Pixel-wise & Mask2Former & 86.0\phantom{ / 00.0$^*$} & 86.1\phantom{ / 00.0$^*$}\\
\textbf{This method} & Boxes & Mask2Former & 86.2\phantom{ / 00.0$^*$}   & \multicolumn{1}{c}{-}\\
\  & Scribbles & Mask2Former & 85.9\phantom{ / 00.0$^*$}  & \multicolumn{1}{c}{-}\\
\  & Boxes, Scribbles & Mask2Former & \textbf{88.6}\phantom{ / 00.0$^*$}  & \textbf{89.1}\phantom{ / 00.0$^*$} \\
\bottomrule
\end{tabular}
\end{table*}

%% file: tables/first_stage_refuge.tex
\begin{table}[h]
    \centering
    \caption{First-stage performance comparison on the REFUGE2 dataset. To isolate the impact of our logic-guided constraints, we compare our weakly supervised MedSAM directly against zero-shot MedSAM with different prompts and finetuned MedSAM with partial cross-entropy over points and pixel-level supervision. For a broader context, we also include other foundation models fine-tuned with dense supervision. All models are evaluated using non-tight bounding-box prompts. }
    \setlength{\tabcolsep}{4pt}
    \begin{tabular}{l l c c}
        \toprule
        Model & Annotations &  mean Dice & mIoU \\
        \midrule
        SAMed~\cite{zhang2023customized} & Pixel-wise  &  87.1 & 77.6 \\
        SAM-Med2D~\cite{cheng2023sam} & Pixel-wise  &  \textbf{89.2} & \textbf{81.8} \\
        SAM-U~\cite{deng2023sam} & Pixel-wise  & 89.1 & 81.7 \\
        \midrule
        MedSAM~\cite{ma2024segment} & Points  & 20.8 & 11.6 \\
        MedSAM~\cite{ma2024segment} & Boxes  & 47.7 & 36.7 \\
        \ + finetuning & Boxes, Points & 78.3 & 65.0  \\
        \textbf{\ + this method} & Boxes, Points  & 87.5 & 77.9 \\
        \ + finetuning & Pixel-wise & 88.7 & 81.3 \\
        \bottomrule
    \end{tabular}
    \label{tab:first_stage_refuge}
\end{table}

%% file: tables/ablations.tex
\begin{table}[t]
\centering
\caption{Effect of removing each constraint on pseudo-label quality for the Pascal VOC train set. As a baseline, we include SAM-huge prompted with bounding boxes.}
\begin{tabular}{l r r}
\toprule
\textbf{Used constraints} & \textbf{mIoU(\%)} & \textbf{$\Delta$mIoU(\%)} \\
\midrule
$\mathcal{F}$ (all constraints) & 94.50 &  \\
$\mathcal{F} \setminus \{\phi_{\textbf{neighborhood}}\}$ & 94.17 & -0.33 \\
$\mathcal{F} \setminus \{ \phi_{\textbf{fill}}\}$ & 93.88 & -0.62 \\
$\mathcal{F} \setminus \{\phi_{\textbf{background}}\}$ & 93.87 & -0.63 \\
$\mathcal{F} \setminus \{ \phi_{\textbf{bbox}}\}$ & 93.84 & -0.66 \\
$\mathcal{F} \setminus \{ \phi_{\textbf{scribbles}}\}$ & 91.73 & -2.77 \\
$\mathcal{F} \setminus \{ \phi_{\textbf{borders}}\}$ & 90.96 & -3.54 \\
\midrule
Baseline SAM & 90.50 & -4.00 \\
\bottomrule
\end{tabular}
\label{tab:constraints_ablation}
\end{table}

%% file: sec/8_conclusions.tex
\section{Conclusion}
This work proposes a neurosymbolic approach to weakly supervised semantic segmentation, combining the reasoning power of differentiable fuzzy logic with the representational strength of foundation models. The framework translates weak annotations and expert domain knowledge into logical constraints, unifying them into a single learning objective. Compared to baselines, this logic-guided fine-tuning produces significantly higher-quality pseudo-labels, narrowing the gap to full supervision. Consequently, our second-stage models achieve state-of-the-art WSSS performance on the Pascal VOC dataset and competitive results on the specialized REFUGE2 benchmark, proving the method's effectiveness \footnote{The source code is available online \url{https://github.com/StefanoColamonaco/Logic-Guided-Segmentation}}.

\paragraph{Limitations \& future work.} While effective, our framework's best performance relies on domain-specific logical constraints, which may require domain expertise. To address cases where human experts are unavailable, recent studies have shown that large language models can successfully serve as an alternative source for this domain knowledge~\cite{gupta2023visprog}. Future work could explore applying this framework to even weaker supervision, such as image-level tags, by designing constraints that propagate sparse information~\cite{shindo2024deisam}. A final limitation of our approach is that the fuzzy loss adds computational overhead during the first-stage fine-tuning (see \cref{app:additional_res}), which improved implementations may mitigate. 

\section{Acknowledgements}
This research received funding from the Flemish Government under the “Onderzoeksprogramma Artifici$\ddot{e}$le Intelligentie (AI) Vlaanderen” programme, the KU Leuven Research Fund (GA No. STG/22/021), the European Research Council (ERC) under the European Union’s Horizon 2020 research and innovation program (Advanced Grant DeepLog No. 101142702), the Flemish research foundation (FWO) projects "Neurosymbolic AI and Constraint Learning" (Project G047124N) and "Relational Concept-Based Models" (GA No. G033625N).

%% file: sec/appendix.tex
\crefalias{section}{appendix}

\section{Additional Explanations and Validations}

\subsection{Conditional Independence Assumption} \label{app:CIA}
We approximate $p(\phi \mid \xvars)$ by introducing conditional independence assumptions.
More specifically, for any formulas $\phi_1$ and $\phi_2$, we assume \[p(\phi_1 \land \phi_2 \mid \xvars) = p(\phi_1 \mid \xvars) \cdot p(\phi_2 \mid \xvars).\] 

This approximation is also known as product t-norm semantics in fuzzy logic~\cite{novak1999mathematical,diligenti2017semantic}.
Next, disjunctions can be reduced to conjunctions and negations using De Morgan's law. So as $p(\neg \phi \mid \xvars) = 1 - p(\phi \mid \xvars)$, it follows that:
\begin{align*}
p(\phi_1 \lor \phi_2 \mid \xvars) &= p(\neg (\neg \phi_1 \land \neg \phi_2) \mid \xvars) \\
&= 1 - p(\neg \phi_1 \land \neg \phi_2 \mid \xvars) \\
&= 1 - p(\neg \phi_1 \mid \xvars) \cdot p(\neg \phi_2 \mid \xvars) \\ 
&= 1 - (1 - p(\phi_1 \mid \xvars)) \cdot (1 - p(\phi_2 \mid \xvars))
\end{align*}

Other logical operations are reducible to conjunctions and disjunctions. For instance, implications can be rewritten as $\phi_1 \Rightarrow \phi_2 \equiv \neg \phi_1 \lor \phi_2$.

Compared to other fuzzy relaxations, the product t-norm has a clear interpretation as the probabilistic semantics with conditional independence assumptions. Furthermore, other
fuzzy operators such as the G\"odel (min/max) t-norm
and the {\L}ukasiewicz t-norm
suffer from large flat regions with no gradient~\cite{vankrieken2022analyzing}.


\subsection{Expressivity and the Connection to Mean-Field CRFs} \label{app:expressivity}
Our constraints subsume several existing structured-regularization
objectives: $\phi_{\textbf{bbox}}$ recovers the multiple-instance
bounding-box prior~\cite{kervadec2020bounding}, while $\phi_{\textbf{borders}}$ express the pairwise smoothness of a Potts
CRF. The advantage of our formulation is not the choice of any single
constraint but the uniform mechanism that compiles all of them. A CRF
couples its potentials through an inference procedure, typically
iterative mean-field, that must be re-derived for each potential family:
pairwise terms admit fast filtering, whereas higher-order terms demand
dedicated constructions ($P^n$-Potts, robust $P^n$, pattern-based)
engineered so that an efficient update remains
available~\cite{krahenbuhl2011efficient,arnab2016higher}. Under the conditional-independence relaxation (\cref{app:CIA}),
$p(\phi \mid \xvars)$ is instead evaluated in closed form by structural
recursion over the formula (\cref{app:diff_imp}), in a single
pass over the network logits and at cost linear in the formula size,
independent of constraint order. Unary, pairwise, and higher-order
constraints are thus optimized through one differentiable relaxation,
with no per-potential inference and no unrolled solver to backpropagate
through.

The connection to $\phi_\textbf{borders}$ goes further: its loss is an upper bound on
the energy that mean-field inference minimizes directly. We write
$p_{i,j} = f_\theta(x)_{i,j} \in \Delta$ for the predicted distribution at pixel
$(i,j)$. Since $f_\theta$ factorizes over pixels, two distinct pixels are independent
given $x$, so the probability that they share a label is exactly the inner product of
their predictions,
\begin{equation}
\nonumber
p(Y_{i,j} = Y_{u,v} \mid \xvars) = \langle p_{i,j}, p_{u,v} \rangle .
\end{equation}

Let $\mathcal{N}$ collect the constrained neighbor pairs (adjacent pixels in the same
superpixel), so that $\phi_\textbf{borders}$ reduces to
$\bigwedge_{((i,j),(u,v)) \in \mathcal{N}} (Y_{i,j} = Y_{u,v})$. In log-space
(~\cref{app:diff_imp}) its loss is the sum of negative log-agreements.
Reading the same predictions as the fully factorized distribution
$Q = \prod_{i,j} p_{i,j}$ gives a unit-weight Potts model the expected pairwise energy
$\bar{E}(Q)$:
\begin{equation}
\nonumber
\mathcal{L}(x, \phi_\textbf{borders}) = -\sum_{\mathcal{N}} \log \langle p_{i,j}, p_{u,v} \rangle,
\qquad
\bar{E}(Q) = \sum_{\mathcal{N}} \bigl( 1 - \langle p_{i,j}, p_{u,v} \rangle \bigr).
\end{equation}
$\bar{E}(Q)$ is the pairwise term that regularized-loss smoothness objectives
minimize. DenseCRF mean-field~\cite{krahenbuhl2011efficient} also carries unary
potentials, an entropy term, and Gaussian edge weights, which $\mathcal{N}$ omits.

\begin{proposition}
$\mathcal{L}(x, \phi_\textbf{borders}) \geq \bar{E}(Q)$, with equality if and only if
$\langle p_{i,j}, p_{u,v} \rangle = 1$ for every pair in $\mathcal{N}$.
\end{proposition}

\begin{proof}
The bound follows from the scalar inequality $-\log a \geq 1 - a$, which holds for all
$a \in (0,1]$ with equality only at $a = 1$. To see this, set $h(a) = -\log a - (1-a)$.
Its derivative $h'(a) = (a-1)/a$ is non-positive on $(0,1]$, so $h$ is non-increasing
there; together with $h(1) = 0$, this gives $h(a) \geq h(1) = 0$, strictly unless
$a = 1$. Each agreement $\langle p_{i,j}, p_{u,v} \rangle$ lies in $(0,1]$ (softmax
predictions are strictly positive), so applying the inequality termwise and summing
over $\mathcal{N}$ yields $\mathcal{L}(x, \phi_\textbf{borders}) \geq \bar{E}(Q)$.
Equality holds only when every term attains $a = 1$, i.e. when all constrained pairs
agree with certainty.
\end{proof}

The two objectives therefore share their minimizers, both vanishing exactly on the
smoothness-satisfying configurations. Since $1 - a$ is the first-order term of
$-\log a$ about $a = 1$, the mean-field energy is the linearization of our loss around
perfect agreement, and the regularized loss of Tang \emph{et al.}~\cite{tang2018regularized}
is recovered as this first-order relaxation. Retaining the full logarithm scales each pair's
gradient by $\langle p_{i,j}, p_{u,v} \rangle^{-1}$, so it sharpens the update on
confidently violated pairs. Optimizing $\phi_\textbf{borders}$ thus behaves like an
end-to-end mean-field CRF without unrolling its iterations, obtained from a single
logical formula rather than a hand-designed potential.

The same equivalence holds for any conjunction of pairwise agreement atoms. This is, however, strictly a pairwise phenomenon. Higher-order constraints such as $\phi_{\textbf{bbox}}$ admit no such
Potts energy, yet are compiled by the same relaxation,
which is precisely where our formulation differs from the CRF.

\subsection{Inexact Constraints}\label{subsec:inexact-constraints}
Logical constraints can sometimes be imperfect. The real object is not to maximize the constraint probability $p(\phi \mid \xvars)$ but the ground truth segmentation $p(\yvars^* \mid \xvars)$. We can view their relation as follows.

\begin{align}
\begin{split}
p(\yvars^* \mid \xvars) &= \sum_{i \in \{\top, \bot\}} p(\phi=i, \yvars^* \mid \xvars) \\
&= \sum_{i \in \{\top, \bot\}} p(\phi=i \mid \xvars) p(\yvars^* \mid \phi=i, \xvars) \\
&= (\alpha - \beta) \cdot p(\phi = \top \mid \xvars) + \beta \\
& \quad \text{ with } \alpha = p(\yvars^* \mid \phi=\top, \xvars) \ \text{ and } \beta = p(\yvars^* \mid \phi=\bot, \xvars) 
\end{split}
\end{align}

In case the constraint is perfect, \ie, only true for the ground truth segmentation map $\yvars^*$ ($\yvars \models \phi \Leftrightarrow \yvars = \yvars^*$), we get $\alpha =1$, $\beta=0$ and hence $p(\yvars^* \mid \xvars) = p(\phi = \top \mid \xvars)$. In case the constraint always holds for the ground truth segmentation map, but also for some other segmentation maps, $\beta = 0$ and $p(\yvars^* \mid \xvars) \propto p(\phi = \top \mid \xvars)$. So $\alpha$ does not depend on the constraint and can be ignored. 

The interesting case is when some constraints are inexact, meaning it is possible that $\yvars^* \not\models \phi$, which makes $\alpha$ and $\beta$ depend on the constraint. This is the approach taken in the paper.
However, in case there is a (small) set of ground truth segmentation maps available, it is also possible to estimate $\alpha$ and $\beta$ on a small validation set for better results.

\subsection{Equivalence of Semantic Loss and Cross-Entropy on Full Supervision} \label{app:equivalence}
We start by expanding the loss function for the fully supervised constraint $\phi_\textbf{fs}$.
\[
\mathcal{L}_{\phi_\textbf{fs}}(x) = -\log p\left( \bigwedge_{i,j} (Y_{i,j}=y_{i,j}^{*}) \,\middle\vert\, \xvars \right)
\]
Applying the conditional independence assumption, we can convert the probability of the conjunction over all pixels into the product of the probabilities for each individual pixel,
\[\mathcal{L}_{\phi_\textbf{fs}}(x) = -\log \left( \prod_{i,j} p(Y_{i,j}=y_{i,j}^{*} \mid \xvars) \right)
\]
Using the logarithmic identity $\log(\prod_i a_i) = \sum_i \log(a_i)$, we transform the product inside the log into a summation.
\[
\mathcal{L}_{\phi_\textbf{fs}}(x) = - \sum_{i,j} \log p(Y_{i,j}=y_{i,j}^{*} \mid \xvars)
\]
The term $p(Y_{i,j}=y_{i,j}^{*} \mid \xvars)$ represents the model's predicted probability that the pixel at $(i,j)$ belongs to the ground-truth class $y_{i,j}^{*}$. Consequently, the expression $-\sum_{i,j} \log p(Y_{i,j}=y_{i,j}^{*} \mid \xvars)$ is the standard definition of pixel-wise cross-entropy loss. Thus, strictly minimizing $\mathcal{L}(\xvars, \phi_\textbf{fs})$ is equivalent to minimizing the cross-entropy loss.

\section{Detailed Method Formulation} \label{app:method_formulation}

\subsection{Weakly Supervised Constraints}

\textbf{Bounding-box tightness.} Each row and each column of the bounding box contains a pixel with the target class $\yvar^*$.
\begin{align}
 \phi_\textbf{bbox} \coloneq \left( \bigwedge_{i \in [i_1..i_2]} \bigvee_{j \in [j_1..j_2]} (\Yvar_{i,j} = \yvar^*) \right) \land \left( \bigwedge_{j \in [j_1..j_2]} \bigvee_{i \in [i_1..i_2]} (\Yvar_{i,j} = \yvar^*) \right) \nonumber    
\end{align}

\noindent \textbf{Background.} Every pixel not contained in any bounding-box ($\mathcal{B}$) must take the background class.

\[
\phi_{\textbf{background}} := \bigwedge_{\substack{i \in [1..n] \\ j \in [1..m]}} \left( \left( \bigwedge_{b \in \mathcal{B}} (i,j) \notin b \right) \Rightarrow (Y_{i,j} = \texttt{background}) \right)
\]

\subsection{Unsupervised Constraints} \label{app:unsup_con}

\paragraph{Smoothness.} Segmentation maps typically satisfy some smoothness constraints. One example is that for each pixel, at least one of the neighbors should have the same label. This avoids isolated artifacts.
\[
\phi_\textbf{neighborhood} \coloneq \bigwedge_{\substack{i \in [1..n] \\ j \in [1..m]}} \bigvee_{(i',j') \in \text{nb}(i,j)} (\Yvar_{i,j} = \Yvar_{i',j'})    
\]
In our implementation, the neighborhood 
nb$(i,j)$ of a pixel $(i,j)$ is defined as the set of its 8 surrounding pixels (the standard Moore neighborhood).

A weaker version of this constraint ($\phi_\textbf{neighborhood} \models \phi_\textbf{fill}$) states that pixels surrounded by pixels with the same class should also have that same class.
\[
\phi_\textbf{fill} \coloneq \bigwedge_{\substack{i \in [1..n]\\j\in[1..m]}} 
\left(
\bigwedge_{\substack{(i_1,j_1) \in \text{nb}(i,j) \\ (i_2, j_2) \in \text{nb}(i,j) }} \Yvar_{i_1,j_1} = \Yvar_{i_2,j_2}  
\right)
\Rightarrow
\left(
\bigwedge_{\substack{(i_1,j_1) \in \text{nb}(i,j) }} \Yvar_{i,j} = \Yvar_{i_1,j_1}  
\right)
\]

\paragraph{Superpixels.} Let $S$ be the superpixel map where $S_{i,j}$ denotes the superpixel index of the pixel at coordinates $(i,j)$. The constraint enforces that if a pixel $(i,j)$ and its neighbor $(u,v)$ belong to the same superpixel (\ie, $S_{i,j} = S_{u,v}$), they must imply the same class prediction ($Y_{i,j} = Y_{u,v}$):
\[
\phi_\textbf{borders} \coloneq \bigwedge_{\substack{i \in [1..n]\\j\in[1..m]}} \bigwedge_{(u,v) \in \text{nb}(i,j)} (S_{i,j} = S_{u,v} \Rightarrow Y_{i,j} = Y_{u,v})
\]
In this formulation, the implication $A \Rightarrow B$ ensures that the constraint is trivially satisfied (true) if the pixels cross a superpixel boundary (where $S_{i,j} \neq S_{u,v}$), thereby allowing label transitions to occur freely at those locations.

\paragraph{Corners.} Let $B$ be a bounding box defined by its vertical and horizontal limits $[y_{\min}..y_{\max}]$ and $[x_{\min}..x_{\max}]$, associated with a target class $y^*$. We define the spatial center of the box as $(c_y, c_x)$ and its radii as $r_y = \frac{y_{\max} - y_{\min}}{2}$ and $r_x = \frac{x_{\max} - x_{\min}}{2}$. The inscribed ellipse within this bounding box is given by the standard equation:
\begin{equation}
    \frac{(i - c_y)^2}{r_y^2} + \frac{(j - c_x)^2}{r_x^2} \leq 1
\end{equation}
We define the set of corner pixels $E_{box}$ as all pixels $(i, j)$ inside $B$ that strictly fall outside this ellipse (\ie, where the left side of the equation is strictly greater than 1). The constraint enforces that these corner pixels must not take the target class label $y^*$.

\begin{equation}
    \phi_\textbf{corners} := \bigwedge_{(i,j) \in E_{box}} \neg(Y_{i,j} = y^*)
\end{equation}

\subsection{Differentiable Implementation} \label{app:diff_imp}

The implementation computes the fuzzy loss in log-space to prevent numerical underflow.
\begin{align*}
    \log p(\phi_1 \land \phi_2) &= \log p(\phi_1) + \log p(\phi_2) \\
    \log p(\phi_1 \lor \phi_2) &= \log [1 - (1 - p(\phi_1))(1-p(\phi_2))]
\end{align*}
As most constraints range over all pixels in the image, we tensorize the computations using PyTorch, allowing efficient GPU-based computation.

\section{Training Details}
\label{app:training_details}

\subsection{Datasets} \label{app:datasets}

\paragraph{Pascal VOC 2012.} The Pascal Visual Object Classes (VOC) 2012 dataset is the standard benchmark for evaluating Weakly Supervised Semantic Segmentation (WSSS) methods. It contains images of 20 foreground object categories along with a background class. The original dataset provides 1,464 images for training, 1,449 for validation, and 1,456 for testing. Following standard practice in WSSS literature, we utilize the augmented training set provided by~\citet{hariharan2011semantic}, which includes additional annotations from the Semantic Boundaries Dataset (SBD), expanding the total number of training images to 10,582. For weak supervision, we utilize the established ground-truth bounding boxes and the scribble annotations.
As noted in the main text, we reserve a 10\% subset of the training data for internal validation during the second-stage training. This internal split allows us to monitor convergence and perform hyperparameter selection without any data leakage from the original Pascal VOC validation set, which we treat as an unseen test set for internal reporting. Finally, to ensure a fair comparison with the state-of-the-art, all reported test set results are obtained by submitting our final model predictions to the official Pascal VOC evaluation server.

\paragraph{REFUGE2.} The REFUGE2 (REtinal FUndus Glaucoma challengE 2) dataset is a specialized medical imaging benchmark designed for glaucoma detection and optic nerve head segmentation. It consists of 1,200 color retinal fundus images, traditionally split into 400 training, 400 validation, and 400 testing images. The dense segmentation task focuses on accurately localizing two distinct anatomical structures: the optic disc and the smaller, contained optic cup. 

Unlike Pascal VOC, REFUGE2 provides only dense, pixel-perfect ground-truth masks. To adapt this dataset for our weakly supervised setting, we discard the dense masks during training and synthetically derive weak annotations (bounding boxes and points) directly from the ground truth:
\begin{itemize}
    \item \textbf{Bounding Boxes:} For each image, we extract tight bounding boxes by identifying the four extreme spatial coordinates (minimum and maximum $x$ and $y$ values) of the ground-truth segmentation regions for both the optic disc and the optic cup. 
    \item \textbf{Point-level Supervision:} To simulate sparse human point clicks or scribbles, we randomly sample three distinct pixel coordinates from within each target class region present in the image, as well as three points from the background region. 
    \item \textbf{Test-Time Prompt Jitter:} During inference, human annotators rarely provide perfect, pixel-tight bounding boxes. To simulate a more realistic and challenging deployment scenario, we evaluate our first-stage model's robustness using non-tight bounding boxes. Specifically, we apply a random spatial jitter to the coordinates of the tight bounding boxes such that the resulting prompt maintains only a 75\% overlap with the target segmented region.
\end{itemize}

\subsection{General Training Details}

\paragraph{Hardware.} We run all our experiments on a Nvidia L40S 48GB GPU card with AMD EPYC 9334 CPU and 256GB RAM.
\paragraph{Total runtime.} The first-stage fine-tuning of the foundation model requires approximately 6 hours for the Pascal VOC dataset and 1.5 hours for the REFUGE2 dataset. Note that these reported training times are approximate; actual runtimes may fluctuate depending on the specific hardware configuration and the concurrent computational load on the machine during execution.
\paragraph{Reproducibility. }To ensure the robustness and reproducibility of our findings, all experiments were repeated using three different random seeds (0, 1, and 2), and the reported quantitative results reflect the average performance across these runs. In \cref{tab:constraints_ablation_stdev,tab:secondstage_pascal_stdev,tab:refuge_stdev} we report the mean and std obtained over the runs. The source code is available online.\footnote{\url{https://github.com/StefanoColamonaco/Logic-Guided-Segmentation}}

\subsection{Stage 1: Logic-Guided SAM Fine-tuning}

\paragraph{Architecture.}
We utilize the Huge variant of the Segment Anything Model (SAM) as our base model for the Pascal dataset, while for the REFUGE2 dataset we utilize the MedSAM-b variant. To maintain computational efficiency and leverage the robust pre-trained representations, we freeze the parameters of both the Image Encoder and the Prompt Encoder. Fine-tuning is restricted exclusively to the Mask Decoder, which learns to interpret the logic-driven supervision signals.

\paragraph{Optimization.}
The model is trained using the Adam optimizer with a learning rate of $1 \times 10^{-4}$ and no weight decay. We employ a physical batch size of 16 images per GPU.

\paragraph{Training Schedule.}
Training proceeds for 30 epochs on both Pascal VOC augmented training set and REFUGE2. Validation is performed at the end of every epoch to monitor constraint satisfaction and segmentation quality. 

\paragraph{Logical Constraints.}
The loss function is defined as the summation of the negative log-likelihoods of the satisfied constraints. Based on the available weak annotations and prior knowledge, we incorporate the following set of constraints.
\begin{itemize}
    \item \textbf{Weak Label Constraints:} $\phi_\textbf{bbox}$ (bounding box tightness), $\phi_\textbf{scribbles}$ (scribble consistency), and $\phi_\textbf{background}$ (background exclusion).
    \item \textbf{Structural Priors:} $\phi_\textbf{neighborhood}$ (neighbor consistency), $\phi_\textbf{fill}$ (region filling), $\phi_\textbf{borders}$ (superpixel-based boundary preservation) and,
    $\phi_\textbf{corners}$ (shape prior, used only in REFUGE2).
\end{itemize}
All constraints are weighted equally in the final loss objective.

\subsection{Stage 2: Fully Supervised Segmentation}
In the second stage, the pseudo-labels generated by the fine-tuned SAM are treated as ground-truth annotations to train standard, prompt-free segmentation networks. We employ two architectures to ensure fair comparison with existing literature:

\paragraph{DeepLabV2.} For our experiments with convolutional architectures, the DeepLabV2 model was instantiated with an output stride of 16. We optimized the network using Stochastic Gradient Descent (SGD) with a momentum of 0.9 and a weight decay of 0.0005. The model was trained for 20 epochs with a total batch size of 10. We employed a base learning rate of 0.00025, which was modulated by a polynomial decay schedule with a power of 0.9. To prevent the loss of pretrained feature representations, we applied a layer-specific learning rate strategy: the backbone received a 0.5$\times$ multiplier (yielding an effective rate of 0.000125), while the ASPP and classifier heads were updated at the base rate. The training objective was the standard Cross-Entropy Loss. Model selection was performed by monitoring the Mean Intersection over Union (mIoU) on the validation subset of the training data. At inference, we utilize a multi-scale evaluation strategy (scales: 0.5, 0.75, 1.0, 1.25, 1.5), fusing the predictions via a maximum response across scales, followed by optional DenseCRF post-processing.

\paragraph{Mask2Former.} For our transformer-based approach, we fine-tuned the Mask2Former architecture using a Swin-Large backbone pretrained on COCO instance segmentation. The training was conducted for 20 epochs using a batch size of 4. We utilized the AdamW optimizer with zero weight decay and a base learning rate of 0.0001. Similar to the DeepLabV2 setup, a polynomial decay schedule (power = 0.9) was applied over the total iterations. To avoid catastrophic forgetting of the pretrained representations, a layer-wise decay was implemented, dividing the backbone’s learning rate by a factor of 10. Input images were resized and normalized using standard ImageNet statistics. As with our other models, final model selection was based on the peak validation mIoU.

\section{Additional Results} \label{app:additional_res}

\textbf{Additional comparisons for finetuned MedSAM:} We implemented additional baselines fine-tuning SAM with partial cross-entropy on scribbles, which yielded significantly lower performance than our logic-guided approach (\cref{tab:baselines}). To further contextualize this, we also evaluated a standard "Self-Training" baseline (finetuning on zero-shot pseudo labels).

\begin{table}[h]
\centering
\caption{REFUGE2 baseline comparison demonstrating that our logic-guided constraints outperform naive partial supervision. }
\label{tab:baselines}
\begin{tabular}{llrr}
\toprule
Method & Annotations & Dice & mIoU  \\
\midrule
MedSAM Self-Training & Zero-shot pseudo-labels  & $49.11\pm 1.03$ & $37.71\pm 0.43$ \\
MedSAM + finetuning & Prompt with boxes + partial CE with points  &$78.35\pm 1.71$ & $65.05\pm 1.57$\\
\textbf{This method} & Boxes + points  & $\mathbf{87.50} \pm 1.63$ & $\mathbf{77.91} \pm 1.66$ \\
\bottomrule
\end{tabular}
\end{table}

\textbf{Constraint satisfaction without fine-tuning:} We evaluated zero-shot SAM constraint satisfaction on the Pascal VOC validation set and found it frequently violates these structural priors, strongly motivating our method. Precisely because these constraints are initially unsatisfied, our framework has the necessary signal to supervise the model; if satisfaction were already 100\%, the logic-guided loss would provide no gradient for structural learning. After fine-tuning, satisfaction rates improve (\cref{tab:constraints_sat}).

\begin{table}[h]
\centering
\caption{Constraint satisfaction rates (\%) on Pascal VOC Validation set.}
\label{tab:constraints_sat}
\begin{tabular}{lrr}
\toprule
Constraint & Zero-shot SAM & This method (Fine-tuned) \\
\midrule
Scribble & 65.40 & $73.75 \pm 0.14$ \\
Neighbours & 46.72 & $74.57 \pm 1.61$ \\
Filling & 53.14 & $85.51 \pm 0.78$ \\
Background & 35.68 & $45.66 \pm 0.15$ \\
BBox & 19.24 & $20.90 \pm 0.54$ \\
\bottomrule
\end{tabular}
\end{table}

\textbf{Additional results on REFUGE2.} \cref{tab:first_stage_refuge_full} reports further results on the first stage in the REFUGE2 dataset, including the Dice score and IoU values for the optical disc/cup classes separately. \cref{tab:second_stage_refuge} reports the structure-wise breakdown of our prompt-free second-stage network (Mask2Former) on REFUGE2, against established architectures trained with full pixel-level supervision. While a gap remains on the optic disc, our weakly supervised model matches or exceeds several fully supervised baselines on the smaller optic cup.

\input{tables/first_stage_refuge_full}

\input{tables/second_stage_refuge}

\textbf{Extended state-of-the-art comparison on Pascal VOC.} For completeness, \cref{tab:first_stage_pascal_full,tab:second_stage_pascal_full} extend \cref{tab:first_stage_pascal,tab:second_stage_pascal} with additional recent methods. \cref{tab:first_stage_pascal_full} reports first-stage pseudo-label quality and \cref{tab:second_stage_pascal_full} the final second-stage segmentation mIoU. Across both stages, our method remains competitive with or ahead of the broader set of state-of-the-art approaches.

\input{tables/first_stage_pascal_full}
\input{tables/second_stage_pascal_full}

\textbf{Results with standard deviations.} \cref{tab:constraints_ablation_stdev,tab:secondstage_pascal_stdev,tab:refuge_stdev} report the mean and standard deviation of our method across multiple random seeds, complementing the main-text results.

\input{tables/results_with_std}

\clearpage
\subsection{Computational Complexity and Runtime Analysis}
In this section, we provide a detailed breakdown of the computational overhead introduced by the semantic-based regularization constraints during the fine-tuning of the Segment Anything Model (SAM).

Integrating logic constraints via semantic regularization involves an additional forward pass through the constraint logic and backpropagation through the resulting loss terms. As shown in the table below, most individual constraints (such as scribbles, background, and neighbours) maintain a runtime performance comparable to the baseline Cross-Entropy training (\cref{tab:runtimes}). The $\mathcal{F}$ (all constraints) configuration, which evaluates all logic constraints simultaneously, represents the upper bound of our computational cost at 0.6387 s/iter. This increase is expected given the cumulative nature of the constraints, yet it remains within a manageable range for standard deep learning hardware. 
It is interesting to observe that the evaluation of certain individual constraints exhibits a runtime comparable to, or even slightly lower than, the baseline cross-entropy loss. This efficiency arises because these logical constraints often bypass the need to load, resize, and spatially align dense ground-truth segmentation masks. Instead, they rely purely on fast, memory-contiguous tensor operations applied directly to the network's output logits.

\begin{table}
\centering
\caption{Average runtime per iteration (including both forward and backward pass) for different logic constraint configurations during SAM fine-tuning on the PASCAL Dataset.}
\begin{tabular}{lrrr}
\toprule
\textbf{Configuration} & \textbf{Runtime (ms/iter)} & \phantom{$***$} &  \textbf{$\Delta$ vs. Baseline (\%)} \\
\midrule
Scribbles              & $346.6 \pm 35.5$     &      & $+1.4\%$  \\
Bboxes                 & $479.5 \pm 68.7$     &      & $+40.3\%$ \\
Background             & $334.1 \pm 35.2$     &      & $-2.2\%$  \\
Neighbours             & $378.5 \pm 54.5$     &      & $+10.8\%$ \\
Fill                   & $378.3 \pm 49.4$     &      & $+10.7\%$ \\
Borders                & $389.8 \pm 40.2$     &      & $+14.1\%$ \\
Corners                & $357.8 \pm 41.0$     &      & $+4.7\%$  \\
\midrule
$\mathcal{F}$ (all constraints) & $638.7 \pm 64.1$  & & $+86.9\%$ \\
Baseline SAM (cross-entropy)    & $341.7 \pm 43.1$  & & $-$       \\
\bottomrule
\end{tabular}
\label{tab:runtimes}
\end{table}

\clearpage
\subsection{More Visualizations}

\begin{figure}[h]
    \centering
    \includegraphics[width=0.85\linewidth]{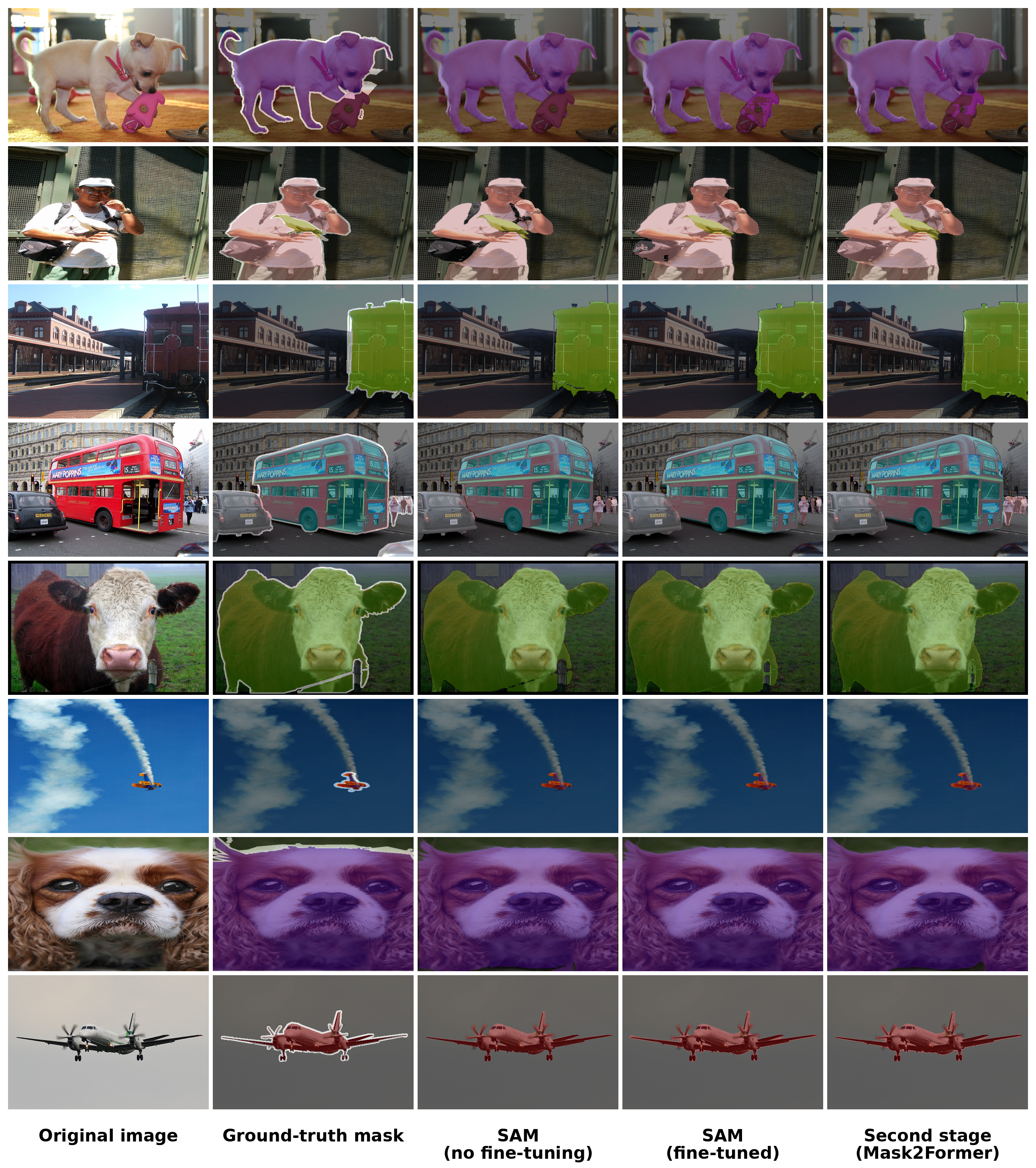}
    \caption{Additional comparison of our two-stage weakly supervised segmentation pipeline on the Pascal VOC 2012 validation set. SAM is prompted with bounding boxes.}
    \label{fig:more_vis_pascal1}
\end{figure}

\begin{figure}[h]
    \centering
    \includegraphics[width=0.85\linewidth]{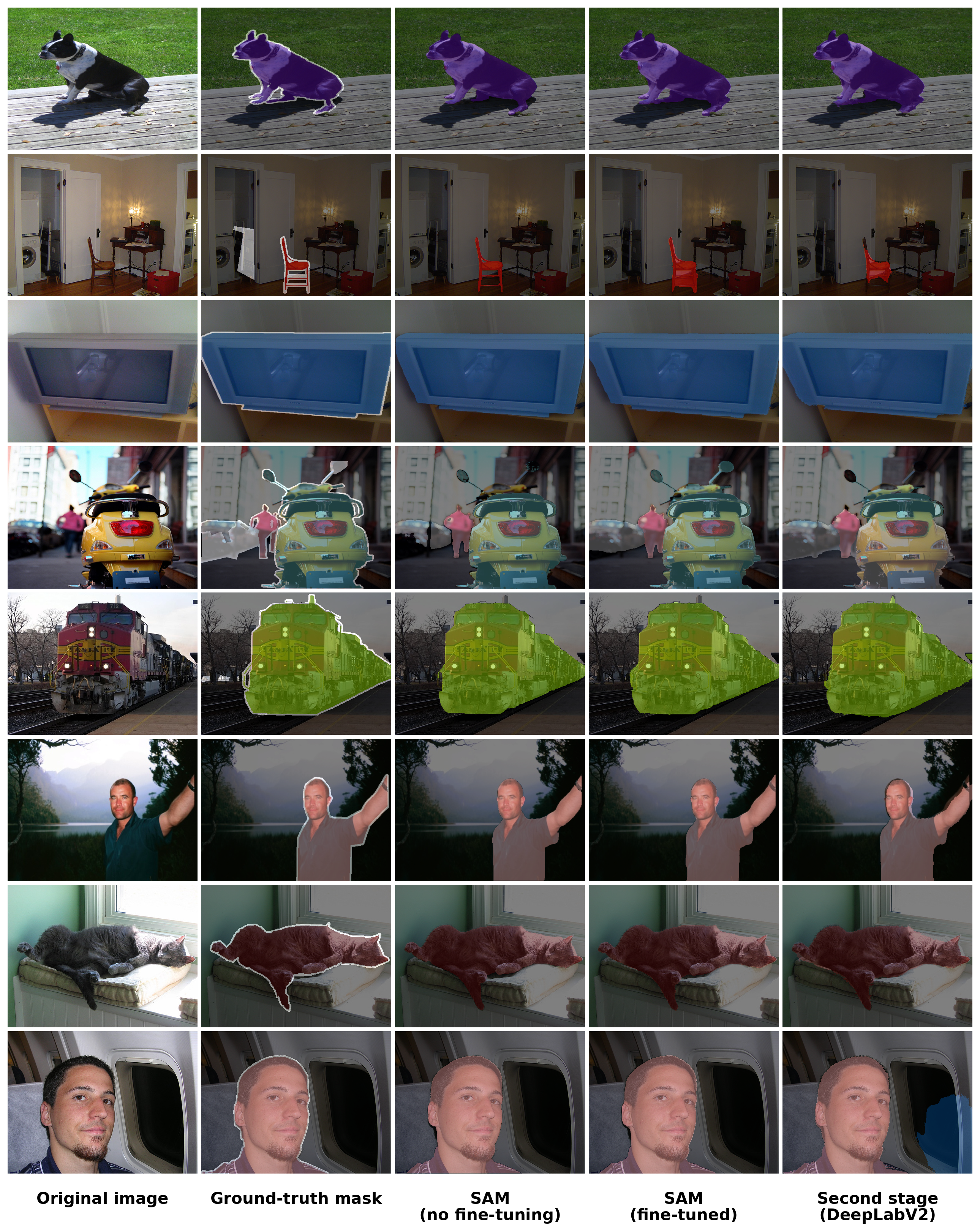}
    \caption{Additional comparison of our two-stage weakly supervised segmentation pipeline on the Pascal VOC 2012 validation set. SAM is prompted with bounding boxes.}
    \label{fig:more_vis_pascal2}
\end{figure}

\begin{figure}
    \centering
    \includegraphics[width=0.7\linewidth]{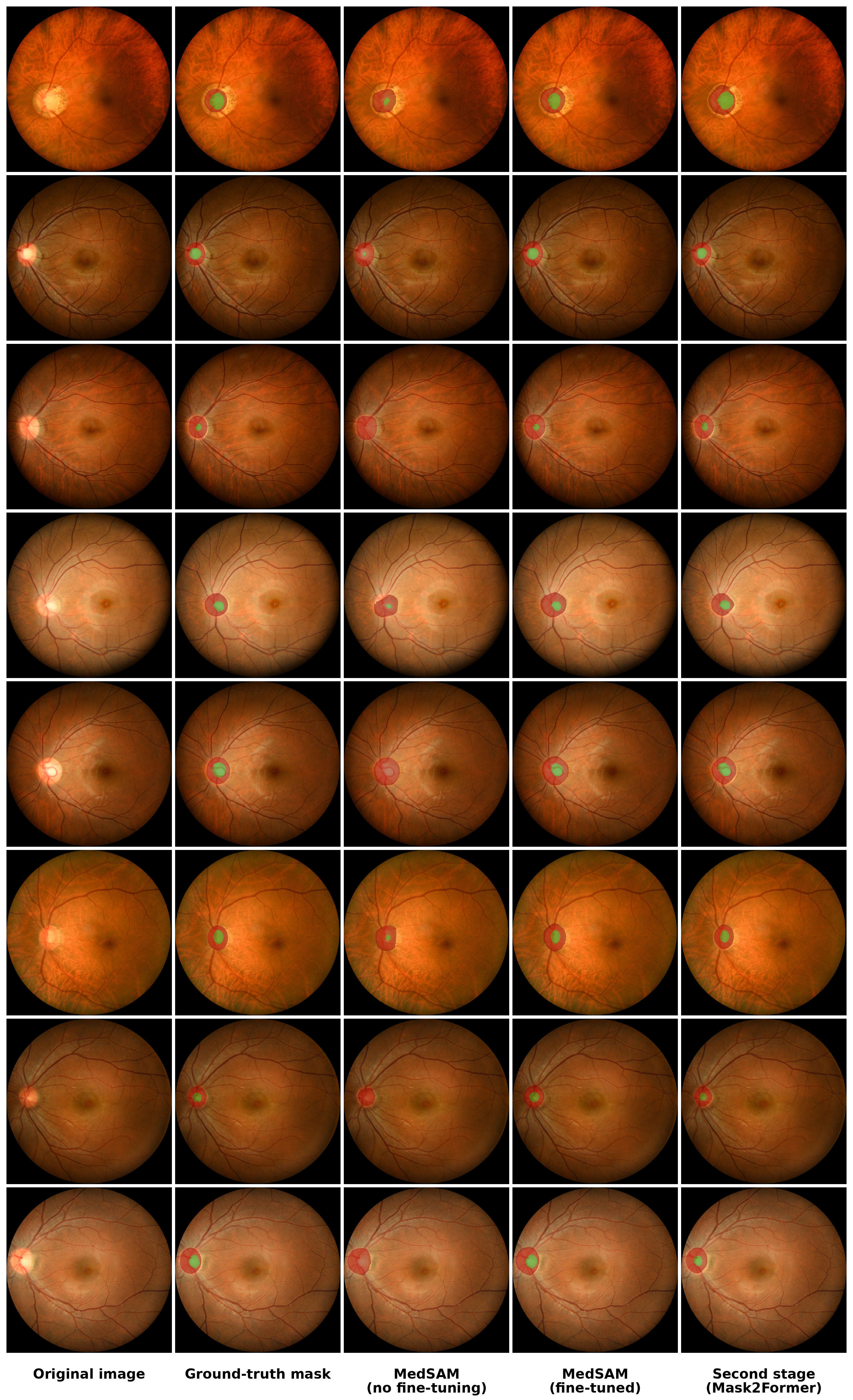}
    \caption{Comparison of our two-stage weakly supervised segmentation pipeline on the REFUGE2 validation set. MedSAM is prompted with bounding boxes.}
    \label{fig:more_vis_refuge2}
\end{figure}

The \cref{fig:more_vis_refuge2,fig:more_vis_pascal1,fig:more_vis_pascal2} show further visualizations of the application of our pipeline on the Pascal VOC and REFUGE2 datasets.


%% file: tables/first_stage_refuge_full.tex
\begin{table*}[h]
    \centering
    \caption{First-stage performance comparison on the REFUGE2 dataset. To isolate the impact of our logic-guided constraints, we compare our weakly supervised MedSAM directly against zero-shot MedSAM with different prompts and finetuned MedSAM with partial cross-entropy over points and pixel-level supervision. For a broader context, we also include other foundation models fine-tuned with dense supervision. All models are evaluated using non-tight bounding-box prompts. }
    \begin{tabular}{l l ccl ccl cc}
        \toprule
        Model & Annotations & \multicolumn{2}{c}{Optic-Disc} & & \multicolumn{2}{c}{Optic-Cup} & & \multicolumn{2}{c}{Mean} \\
        \cmidrule{3-4} \cmidrule{6-7} \cmidrule{9-10} &
        & Dice & IoU & & Dice & IoU & & Dice & IoU \\
        \midrule
        SAMed~\cite{zhang2023customized} & Pixel-wise  & 91.8 & 82.7 & & 82.4 & 72.5 & & 87.1 & 77.6 \\
        SAM-Med2D~\cite{cheng2023sam} & Pixel-wise  & 94.8 & 87.0 & & 83.7 & \textbf{76.6} & & \textbf{89.2} & \textbf{81.8} \\
        SAM-U~\cite{deng2023sam} & Pixel-wise  & \textbf{95.2} & \textbf{88.6} & & 83.0 & 74.9 & & 89.1 & 81.7 \\
        \midrule
        MedSAM~\cite{ma2024segment} & Points  & 21.2 & 11.8 & & 20.4 & 11.4 & & 20.8 & 11.6 \\
        MedSAM~\cite{ma2024segment} & Boxes  & 78.1 & 64.0 & & 17.3 & 9.5 & & 47.7 & 36.7 \\
        MedSAM + partial CE & Boxes, Points& 71.1 & 55.1 && 85.5 & 74.9 && 78.3 & 65.0  \\
        \textbf{This method} & Boxes, Points  & 89.4 & 80.8 & & \textbf{85.7} & 75.0 & & 87.5 & 77.9 \\
        MedSAM~\cite{ma2024segment} & Pixel-wise  & 94.6 & 86.7 & & 82.8 & 75.9 & & 88.7 & 81.3 \\
        \bottomrule
    \end{tabular}
    \label{tab:first_stage_refuge_full}
\end{table*}

%% file: tables/second_stage_refuge.tex

\begin{table*}[h]
    \centering
    \caption{Final segmentation results on the REFUGE2 dataset. We compare the performance of our prompt-free second-stage network (Mask2Former), trained entirely on our generated pseudo-labels, against established medical segmentation architectures trained with full supervision.}
    \begin{tabular}{l l ccl ccl cc}
        \toprule
        Model & Annotations & \multicolumn{2}{c}{Optic-Disc} & & \multicolumn{2}{c}{Optic-Cup} & & \multicolumn{2}{c}{Mean} \\
        \cmidrule{3-4} \cmidrule{6-7} \cmidrule{9-10} & 
        & Dice & IoU & & Dice & IoU & & Dice & IoU \\
        \midrule
             
        ResUNet~\cite{yu2019resunet} & Pixel-wise  & 92.9 & 85.5 & & 80.1 & 72.3 & & 86.5 & 78.9\\
        BEAL~\cite{wang2019beal} & Pixel-wise  & 93.7 & 86.1 & & 83.5 & 74.1 & & 88.6 & 80.1\\
        SegDiff~\cite{amit2021segdiff} & Pixel-wise  & 92.6 & 85.2 & & 82.5 & 71.9 & & 87.5 & 78.5\\
        nnUNet~\cite{isensee2021nnunet} & Pixel-wise  & 94.7 & 87.3 & & 84.9 & 75.1 & & 89.8 & 81.2\\
        TransUNet~\cite{chen2021transunet} & Pixel-wise  & \textbf{95.0} & \textbf{87.7} & & 85.6 & \textbf{75.9} & & \textbf{90.3} & \textbf{81.8}\\
        \midrule
        \textbf{This method} & Boxes, Points  & 88.1 & 78.7 & & \textbf{86.0} & 74.6 & & 87.1 & 76.6 \\
        \bottomrule
    \end{tabular}
    \label{tab:second_stage_refuge}
\end{table*}

%% file: tables/first_stage_pascal_full.tex
\begin{table}[t]
\centering
\caption{Quantitative evaluation of first-stage pseudo-label quality on the Pascal VOC 2012 train set. This table extends Tab.~\ref{tab:first_stage_pascal} with additional state-of-the-art methods.}
\begin{tabular}{l l r}
\toprule
\textbf{Method} & \textbf{Annotations} & \textbf{mIoU\textsubscript{PL}(\%)}\\
\midrule
Full supervision & Pixel-wise & 100.0\\
\midrule
Box2TagBack~\cite{ji2021weakly} & Boxes & 90.2\\
Sun \etal~\cite{sun2023alternative} & ILT & 88.3\\
FMA-WSSS~\cite{yang2024foundation} & ILT & 80.4\\
DHR~\cite{jo2024dhr} & ILT & 83.9\\
SemPLeS~\cite{lin2025semantic} & ILT & 78.4\\
CLIP-ES~\cite{lin2023clip} & ILT, Language & 75.0\\
CLIP-CPAL~\cite{tang2024hunting} & ILT, Language & 75.8\\
VPL~\cite{xu2025toward} & ILT, Language & 80.1\\
\midrule
SAM~\cite{kirillov2023segment,jiang2023segment} & ILT & 61.9\\
 & Points & 71.7\\
 & Scribbles & 89.7\\
 & Boxes & 91.5\\
\quad + combined prompt & Boxes, Scribbles & 91.8\\
\quad + CE finetune & Boxes, Scribbles & 88.4\\
\midrule
\textbf{This method} & Boxes, Scribbles & \textbf{94.5}\\
 & Scribbles & 93.4\\
 & Boxes & 91.7\\
\bottomrule
\end{tabular}
\label{tab:first_stage_pascal_full}
\end{table}

%% file: tables/second_stage_pascal_full.tex
\begin{table*}[t]
\centering
\caption{Quantitative comparison of final second-stage segmentation results on the Pascal VOC 2012 validation and test set. Quantitative comparison of final second-stage segmentation results on the Pascal VOC 2012 validation and test set. In this Table we include more methods from the state-of-the-art.}
\label{tab:second_stage_pascal_full}
\begin{tabular}{l l l r r}
\toprule
\textbf{Method} & \textbf{Annotations} & \textbf{Network} & \textbf{mIoU\textsubscript{val}(\%)} & \textbf{mIoU\textsubscript{test} (\%)} \\
\midrule
TEL~\cite{liang2022tree} & Scribbles & DeepLabV2$^\dagger$ & 77.1\phantom{ / 00.0$^*$} & 74.8 / 76.0$^*$\\
AGMM~\cite{wu2023sparsely} & Scribbles & DeepLabV3+  & 76.4\phantom{ / 00.0$^*$} & \multicolumn{1}{c}{-}\\
Box2TagBack~\cite{ji2021weakly} & Boxes & DeepLabV2 & 75.7\phantom{ / 00.0$^*$} & 75.5\phantom{ / 00.0$^*$} \\
Sun \etal~\cite{sun2023alternative} & ILT & DeepLabV2 & 77.2$^*$ & 77.1$^*$\\
SemPLeS \cite{lin2025semantic} & ILT & DeepLabV2 & 73.9\phantom{ / 00.0$^*$} & 73.8\phantom{ / 00.0$^*$} \\
CLIP-ES~\cite{lin2023clip} & ILT, Language & DeepLabV2 & 73.8\phantom{ / 00.0$^*$} & 73.9\phantom{ / 00.0$^*$}\\
CLIP-CPAL~\cite{tang2024hunting} & ILT, Language & DeepLabV2 & 74.5\phantom{ / 00.0$^*$} & 74.7\phantom{ / 00.0$^*$}\\
Jiang \etal
~\cite{jiang2023segment} & ILT & DeepLabV2 & 71.1\phantom{ / 00.0$^*$} & 72.2\phantom{ / 00.0$^*$}\\
\  & Points & DeepLabV2 & 69.0\phantom{ / 00.0$^*$} & 68.7\phantom{ / 00.0$^*$}\\
\  & Scribbles & DeepLabV2 & 75.9\phantom{ / 00.0$^*$} & 76.6\phantom{ / 00.0$^*$}\\
\  & Boxes & DeepLabV2 & 76.3\phantom{ / 00.0$^*$} & 75.8\phantom{ / 00.0$^*$}\\
Combined prompt & Boxes, Scribbles & DeepLabV2 & 76.1 / 76.7$^*$ & \multicolumn{1}{c}{-} \\ 
Full supervision & Pixel-wise & DeepLabV2 & 76.4 / 77.7$^*$ & \textbf{79.7$^*$} \\
\textbf{This method} & Boxes & DeepLabV2 & 76.4 / 77.4$^*$  & \multicolumn{1}{c}{-}\\
\ & Scribbles & DeepLabV2 & 77.4 / 78.5$^*$  & \multicolumn{1}{c}{-}\\
\  & Boxes, Scribbles & DeepLabV2 & 78.2 / \textbf{79.7$^*$}  & 77.0 / 78.5$^*$\\
\midrule
VPL~\cite{xu2025toward} & ILT, Language & CLIP & 79.3$^*$ & 79.0$^*$\\
CoSA-MS \cite{yang2024weakly} & ILT & UperNet-Swin & 81.4\phantom{ / 00.0$^*$} & 78.4\phantom{ / 00.0$^*$}\\
WeakTr \cite{zhu2023weaktr} & ILT & ViT-S & 78.4\phantom{ / 00.0$^*$} & 79.0\phantom{ / 00.0$^*$}\\
\midrule
FMA-WSSS \cite{yang2024foundation} 
& ILT & Mask2Former & 82.6\phantom{ / 00.0$^*$} & 81.6\phantom{ / 00.0$^*$}\\
DHR \cite{jo2024dhr} 
& ILT & Mask2Former & 82.3\phantom{ / 00.0$^*$} & 82.3\phantom{ / 00.0$^*$}\\
SemPLeS \cite{lin2025semantic} & ILT & Mask2Former & 83.4\phantom{ / 00.0$^*$} & 82.9\phantom{ / 00.0$^*$} \\
Combined prompt & Boxes, Scribbles & Mask2Former & 85.2\phantom{ / 00.0$^*$} & \multicolumn{1}{c}{-} \\ 
Full supervision & Pixel-wise & Mask2Former & 86.0\phantom{ / 00.0$^*$} & 86.1\phantom{ / 00.0$^*$}\\
\textbf{This method} & Boxes & Mask2Former & 86.2\phantom{ / 00.0$^*$}   & \multicolumn{1}{c}{-}\\
\  & Scribbles & Mask2Former & 85.9\phantom{ / 00.0$^*$}  & \multicolumn{1}{c}{-}\\
\  & Boxes, Scribbles & Mask2Former & \textbf{88.6}\phantom{ / 00.0$^*$}  & \textbf{89.1}\phantom{ / 00.0$^*$}\\
\bottomrule
\end{tabular}
\end{table*}

%% file: tables/results_with_std.tex
\begin{table}[h]
\centering
\caption{Quantitative evaluation of first-stage pseudo-label quality on the Pascal VOC 2012 training set. Results report the mean Intersection over Union alongside standard deviations across different logical constraint configurations.}
\begin{tabular}{l r}
\toprule
\textbf{Used constraints} & \textbf{mIoU\textsubscript{PL}(\%)}\\
\midrule
$\mathcal{F}$ (all constraints) & $94.50\pm 0.02$ \\
$\mathcal{F} \setminus \{\phi_{\textbf{neighborhood}}\}$ & $94.17\pm 0.04$  \\
$\mathcal{F} \setminus \{ \phi_{\textbf{fill}}\}$ & $93.88\pm 0.15$ \\
$\mathcal{F} \setminus \{\phi_{\textbf{background}}\}$ & $93.87\pm 0.08$\\
$\mathcal{F} \setminus \{ \phi_{\textbf{bbox}}\}$ & $93.84\pm 0.07$ \\
$\mathcal{F} \setminus \{ \phi_{\textbf{scribbles}}\}$ & $91.73\pm 0.28$\\
$\mathcal{F} \setminus \{ \phi_{\textbf{borders}}\}$ & $90.96\pm 0.23$\\
\bottomrule
\end{tabular}
\label{tab:constraints_ablation_stdev}
\end{table}

\begin{table}[h]
\centering
\caption{Quantitative evaluation of the final second-stage segmentation networks on the Pascal VOC 2012 dataset. Performance is reported as mean Intersection over Union alongside standard deviations across multiple runs.}
\begin{tabular}{l l r}
\toprule
\textbf{Architecture} & Annotations &\textbf{mIoU\textsubscript{val}(\%)}\\
\midrule
DeepLabV2 & Boxes & $76.4\pm 0.02$ \\
DeepLabV2 + CRF & Boxes & $77.4\pm 0.12$ \\
DeepLabV2 & Scribbles & $77.4\pm 0.14$ \\
DeepLabV2 + CRF & Scribbles & $78.5\pm 0.10$ \\
DeepLabV2 & Boxes, Scribbles & $78.2\pm 0.24$ \\
DeepLabV2 + CRF & Boxes, Scribbles & $79.7\pm 0.28$ \\
\midrule
Mask2Former & Boxes & $86.2\pm 0.42$ \\
Mask2Former & Scribbles & $85.9\pm 0.9$ \\
Mask2Former & Boxes, Scribbles & $88.6\pm 0.07$ \\
\bottomrule
\end{tabular}
\label{tab:secondstage_pascal_stdev}
\end{table}

\begin{table}[h]
\centering
\caption{Quantitative evaluation on the REFUGE2 dataset, detailing the performance of both the first-stage fine-tuned MedSAM and the final second-stage segmentation network. Results are reported alongside standard deviations across multiple runs.}
\begin{tabular}{l r r r}
\toprule
\textbf{Architecture} & \textbf{IoU\textsubscript{disc}(\%)} &  \textbf{IoU\textsubscript{cup}(\%)} & \textbf{mIoU(\%)}\\
\midrule
Finetuned MedSAM & $80.83\pm 1.02$ & $75.00\pm 4.12$ & $77.91\pm 1.66$\\
Mask2former & $78.67\pm 1.29$ & $74.64\pm 1.63$ & $76.65\pm 0.86$\\
\bottomrule
\end{tabular}
\label{tab:refuge_stdev}
\end{table}